\documentclass[runningheads]{llncs}
\usepackage{graphicx}
\usepackage{tikz}
\usepackage{comment}
\usepackage{amsmath,amssymb} %
\usepackage{color}
\usepackage{booktabs}
\usepackage{balance}
\usepackage{multirow}
\usepackage{multicol}
\usepackage{mwe}
\usepackage{algorithmic,float}
\usepackage[linesnumbered,ruled,vlined,noline]{algorithm2e}
\usepackage{etoolbox}
\usepackage{relsize}
\usepackage{comment}
\usepackage{enumitem}
\usepackage{adjustbox}
\usepackage[capitalize]{cleveref}
\usepackage{mathtools}
\usepackage{colortbl}
\usepackage{cite}
\usepackage{wrapfig}

\newcommand{\modd}{\mathbin{\mathchoice
  {\textstyle\mathsmaller{\%}}
  {\mathsmaller{\%}}
  {\mathsmaller{\%}}
  {\mathsmaller{\%}}
}}
\usepackage[accsupp]{axessibility}  %

\begin{document}
\pagestyle{headings}
\mainmatter
\def\ECCVSubNumber{4920}  %

\title{Putting 3D Spatially Sparse Networks on a Diet}

\titlerunning{Preprint.}
\author{Junha Lee\inst{1} \and
Christopher Choy\inst{2} \and
Animashree Anandkumar\inst{2,3} \and
Jaesik Park\inst{1}}
\authorrunning{Preprint.}
\institute{${}^1$POSTECH \hspace{2mm} ${}^2$NVIDIA \hspace{2mm} ${}^3$Caltech}

\maketitle

\crefname{section}{Sec.}{Secs.}
\Crefname{section}{Section}{Sections}
\Crefname{table}{Table}{Tables}
\crefname{table}{Tab.}{Tabs.}

\newcommand{\Eq}[1]  {Eq.\ \ref{eq:#1}}
\newcommand{\Eqs}[1] {Eqs.\ \ref{eq:#1}}
\newcommand{\Fig}[1] {Figure \ref{fig:#1}}
\newcommand{\Figs}[1]{Figures \ref{fig:#1}}
\newcommand{\Tbl}[1]  {Table \ref{tbl:#1}}
\newcommand{\Tbls}[1] {Tables \ref{tbl:#1}}
\newcommand{\Sec}[1] {Section \ref{sec:#1}}
\newcommand{\Secs}[1] {Sections \ref{sec:#1}}
\newcommand{\App}[1] {Appendix \ref{app:#1}}
\newcommand{\Alg}[1] {Algorithm \ref{alg:#1}}
\newcommand{\Etal}   {et al.}

\newcommand{\OursAcronym}{WS${}^3$-ConvNet}

\newcommand{\jaesik}[1]{{\textcolor{cyan}{Jaesik: #1}}}
\definecolor{purp}{rgb}{0.65, 0.16, 0.65}
\newcommand{\junha}[1]{{\textcolor{orange}{Junha: #1}}}
\newcommand{\chris}[1]{{\textcolor{blue}{Chris: #1}}}
\newcommand{\todo}[1]{{\textcolor{red}{#1}}}

\newcommand{\minus}[1]{{\textcolor{purple}{$\downarrow$ #1}}}
\newcommand{\plus}[1]{{\textcolor{teal}{$\uparrow$ #1}}}
\newcommand{\nminus}[1]{{\textcolor{teal}{$\downarrow$ #1}}}
\newcommand{\nplus}[1]{{\textcolor{purple}{$\uparrow$ #1}}}

\newcommand{\aminus}[1]{{\textcolor{purple}{#1}}}
\newcommand{\aplus}[1]{{\textcolor{teal}{#1}}}
\newcommand{\anminus}[1]{{\textcolor{teal}{#1}}}
\newcommand{\anplus}[1]{{\textcolor{purple}{#1}}}
\newcommand{\mathMesh}{\mathcal{M}}
\newcommand{\mathFace}{f}
\DeclarePairedDelimiter\ceil{\lceil}{\rceil}
\DeclarePairedDelimiter\floor{\lfloor}{\rfloor}

\setlength{\textfloatsep}{0.8cm}
\setlength{\floatsep}{0.8cm}

\begin{abstract}
\vspace{-2mm}
3D neural networks have become prevalent for many 3D vision tasks including object detection, segmentation, registration, and various perception tasks for 3D inputs. However, due to the sparsity and irregularity of 3D data, custom 3D operators or network designs have been the primary focus of research, while the size of networks or efficacy of parameters has been overlooked. In this work, we perform the first comprehensive study on the weight sparsity of spatially sparse 3D convolutional networks and propose a compact weight-sparse and spatially sparse 3D convnet (\OursAcronym{}) for semantic and instance segmentation on the real-world indoor and outdoor datasets. 
We employ various network pruning strategies to find compact networks and show our \OursAcronym{} achieves minimal loss in performance (2.15\% drop) with orders-of-magnitude smaller number of parameters (99\% compression rate) and computational cost (95\% reduction).
Finally, we systematically analyze the compression patterns of \OursAcronym{} and show interesting emerging sparsity patterns common in our compressed networks to further speed up inference (45\% faster).

\end{abstract}

\section{Introduction}
\label{sec:introduction}
Recent advances in 3D neural networks made various 3D vision applications such as 3D shape classification~\cite{qi2017pointnet,qi2017pointnet++}, semantic segmentation~\cite{choy20194d}, object detection~\cite{qi2019deep,gwak2020generative}, reconstruction~\cite{mescheder2019occupancy}, 3D registration~\cite{choy2019fully,choy2020deep}, and many other tasks accessible.
These work focus on broadening the spectrum of the 3D neural networks, but, over the years, the majority of research on 3D perception focus more on introducing customized convolution kernels that preserve the continuity of point cloud~\cite{qi2017pointnet,qi2017pointnet++,thomas2019kpconv,wang2019dynamic,tatarchenko2018tangent,mao2019interpolated,zhao2021point}. However, these continuous convolutions sometimes show lower performance or slower processing time on large scale datasets such as ScanNet~\cite{dai2017scannet} or autonomous driving datasets~\cite{behley2021ijrr,sun2020scalability} due to the slow and memory intensive nearest neighbor search in the coordinate space.
Specifically, to accelerate the nearest neighbor search on GPU, they use pairwise distance computation that is expensive and requires quadratic $\mathcal{O}(N^2)$ memory footprint.
This limits the resolution of the point cloud and prevents wider adoption on large-scale 3D perception.

Discretized convolutions, on the other hand, can process millions of points with faster inference using the spatially sparse representation and GPU hash tables~\cite{graham2014spatially,choy20194d,gwak2020generative} at the cost of small quantization error ($\sim$1cm). Thus, they show promising results on many large-scale indoor~\cite{armeni_cvpr16,dai2017scannet} and outdoor perception tasks~\cite{behley2021ijrr,sun2020scalability}.
These discrete convolutions are effective in processing large-scale point clouds, yet these networks are still bulky due to their 3D convolution kernels, preventing wide adoption on edge devices, commodity servers, and low-power devices.

In the image domain, we observe significant advances in network pruning techniques that compress a network by removing redundant parameters~\cite{lecun1990optimal,frankle2018the,frankle2019stabilizing,lee2018snip,molchanov2016pruning,mozer1989skeletonization} and network architecture design methods that reduce the memory and computation cost of a network by introducing efficient architectures~\cite{howard2017mobilenets,zhang2018shufflenet,tan2019efficientnet}. Compared with more mature research in 2D network compression, the network pruning and compression of 3D convnets have rarely been studied except for pruning spatially dense 3D convolution kernels for 2D videos~\cite{wang2020pruning3dfilters}.
\begin{figure*}[t!]
\begin{center}
    \centering
     \centering
     \begin{minipage}{0.30\linewidth}
     \includegraphics[width=\textwidth]{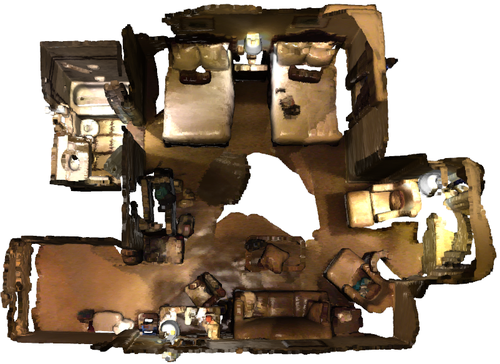}
     \centering\small{(a) Input point cloud}\vspace{8mm}
     \end{minipage}\hfill
     \begin{minipage}{0.30\linewidth}
     \includegraphics[width=\textwidth]{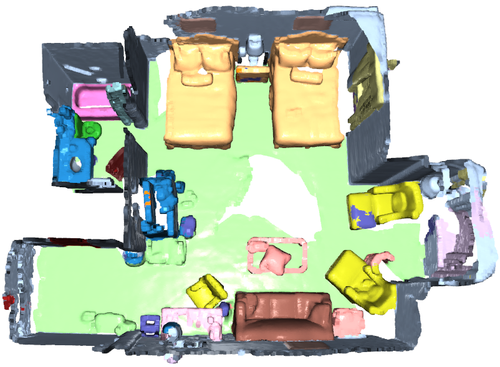}
     \centering\small{(b) Reference Network, \\37.85M Param., 71.57 mIoU(\%)}\vspace{1mm}
     \end{minipage}\hfill
     \begin{minipage}{0.30\linewidth}
     \includegraphics[width=\textwidth]{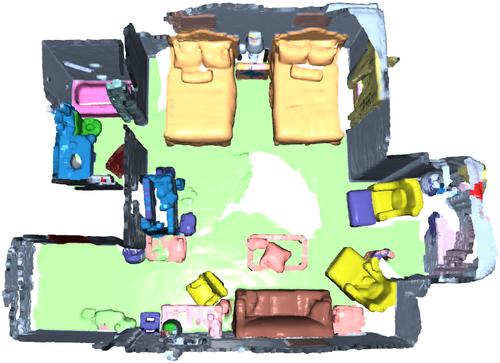}
     \centering\small{(c) Our \OursAcronym, \\\textcolor{red}{0.396M} Param., \textcolor{red}{69.42} mIoU(\%)}
     \end{minipage}\hfill
     \vspace{-2mm}
     \caption{
     Visualization of semantic label prediction of reference neural network (middle) and our \OursAcronym{} that is obtained by pruning \textcolor{red}{99\%} of the weights from the reference neural network (right). While the pruned model has 100 times smaller parameters, the mIoU over the entire ScanNet~\cite{dai2017scannet} validation split is 69.42\%, only the 2.15\% drops}
     \label{fig:teaser}
\end{center}
\end{figure*}

{\bf Our approach:} in this work, to the best of our knowledge, we conduct the first comprehensive study on weight pruning of spatially sparse 3D convolutional networks.
The most effective networks we found show 99\% compression rate with a negligible drop in performance (2.15\% drop) on 3D segmentation tasks. We name these \textit{Weight-Sparse and Spatially Sparse 3D convolutional} networks \textbf{\OursAcronym{}}.
To find these compact 3D networks, we extensively employ the network pruning methods~\cite{han2015learning,lee2018snip,frankle2018the,molchanov2016pruning} on two indoor datasets, ScanNet~\cite{dai2017scannet} and S3DIS~\cite{armeni_cvpr16}, and one outdoor dataset, SemanticKITTI~\cite{behley2019iccv}. 
For indoor datasets, we prune networks for 1) semantic segmentation and 2) instance segmentation creating four distinct tasks for network compression. 
To further validate that our findings can be transferred to more challenging outdoor dataset, we prune semantic segmentation networks on SemanticKITTI~\cite{behley2019iccv} dataset and compare with the state-of-the-art LiDAR-based semantic segmentation methods~\cite{tang2020searching}.

In addition, we propose the weight-sparse spatially sparse convolution (WS${}^3$-conv) algorithm. Network compression results in smaller networks and reduction in memory footprint, but faster inference requires a specialized custom implementation. Our WS${}^3$-conv algorithm shows up to 30\% reduction in inference time.
Lastly, we analyze compression patterns of \OursAcronym{} and propose additional structured pruning on top of the most performant network to further speed up inference.
In summary, our contributions are:

\begin{itemize}[noitemsep,topsep=0pt,parsep=0pt,partopsep=0pt,leftmargin=*]
\renewcommand{\labelitemi}{$\bullet$}
    \item We apply network pruning methods on various 3D segmentation networks trained on various real-world 3D datasets and propose \OursAcronym{}. It achieves 99\% parameter reduction while incurring only 2.15\% drop in accuracy.
    \item We propose an efficient feed-forward algorithm for our \textit{weight-sparse spatially sparse convolution} (WS${}^3$-conv), which enables 30\% faster inference.
    \item Analysis of the pruned spatially sparse convnets reveal interesting properties of 3D convnets. 1) Pruning a larger 3D convnet is more effective. 2) The weight parameters of decoder part of a 3D U-net is less important than the encoder. 3) the pruning patterns reveal that a few spatial convolution kernels have no significance.
    \item We make use of these observations to apply structured pruning on convolution kernels and further speed up our \OursAcronym{}, results in 45\% faster inference while incurring 2.4\% accuracy drop.
\end{itemize}

\section{Related Work}

\subsection{Deep neural networks for 3D data}
We limit the scope of our work to spatial 3D data, although many refer to videos as 3D data, as spatial 3D data requires special operators to handle sparsity and irregularity.
MLP-based methods~\cite{qi2017pointnet,qi2017pointnet++,zhao2021point} directly process the continuous 3D coordinates of point clouds from 3D sensors using a set of MLPs. 
Recent approaches~\cite{mao2019interpolated,thomas2019kpconv,wang2019dynamic,tatarchenko2018tangent} define customized convolution kernels that apply a series of local transformations on each point.
Sparse voxel-based approaches~\cite{graham2014spatially,choy20194d}, on the other hand, discretize point clouds into discrete voxels and apply spatially sparse convolution, which is mathematically equivalent to regular dense convolution.

Each group targets different aspects of point cloud perception and there are pros and cons of each category.
For instance, MLP-based methods~\cite{qi2017pointnet,qi2017pointnet++} are the fastest and the simplest, but often fail to capture local context; others preserve the continuity of points while requiring expensive nearest neighbor search with quadratic memory complexity on GPU~\cite{qi2017pointnet++,thomas2019kpconv,zhao2021point}. 
Voxel-based methods introduce a small quantization error, but discretization provides $\mathcal{O}(1)$ search speed and $\mathcal{O}(N)$ memory footprint using GPU hash tables. Also, thanks to the mathematical equivalence to 2D convnets, we could borrow many successful network architectures such as skip-connection~\cite{he2016deep}, batch normalization~\cite{ioffe2015batch}, and U-shaped network~\cite{ronneberger2015u} from 2D CNNs to 3D spatially sparse CNNs.
Recent study~\cite{tang2020searching} has proposed hybrid architectures that are equipped with both voxel-based and point-based branches and has shown the state-of-the-art performance in 3D outdoor semantic segmentation task. 
They further investigate parameter- and computation-efficient architecture by incorporating neural architecture search (NAS), which makes it, to the best our knowledge, the most relevant prior research to ours.

\subsection{Network compression}
Network pruning or network compression is a technique that removes a portion of weights from a neural network to shrink the size of a network and to reduce the amount of computation required for a forward pass. There are three broad categories of network compression depending on when, where, and how a network is pruned.

\vspace{-4mm}
\subsubsection{Unstructured pruning} removes network weights without a predefined structure.
LeCun~\Etal~\cite{lecun1990optimal} identifies the redundant parameters by analyzing sensitivities of parameters using the second-order derivatives of the objective function.
Later work~\cite{hassibi1993optimal,molchanov2016pruning} replaced the hessian with the Taylor expansion to relax the computation burden of computing the second-order derivatives.
Han~\Etal~\cite{han2015learning} showed that pruning weights with small magnitude and fine-tuning the pruned network leads to a compact model without incurring large loss in accuracy.

\vspace{-4mm}
\subsubsection{Structured pruning} removes a set of network weights with a specific structure (e.g., prune channels, filters altogether) that could provide gains in computation and inference speed without special hardware or software support.
Li~\Etal~\cite{li2016pruning} applied magnitude-based pruning to network filters. Other work~\cite{liu2017learning,huang2018data} learn a scaling factor for each channel and prune an entire channel with a lower magnitude scaling factor.
Other approaches~\cite{luo2017thinet,he2017channel} find channels that don't contribute to the feature reconstruction.

\vspace{-4mm}
\subsubsection{Pruning at initialization}
is a recent network pruning method based on the Lottery Ticket Hypothesis (LTH)~\cite{frankle2018the}, which claims that there is a subnetwork within a network that achieves optimal performance and one can find the subnetwork by rewinding parameters that survive pruning to their initial values.
Frankle~\Etal~\cite{frankle2019stabilizing} use larger architectures by relaxing the restriction of reverting the weights to initial values. 
Zhou~\Etal~\cite{zhou2019deconstructing} show that the difference between the initial weights and fine-tuned weights can be another pruning criterion.
The proxies for determining lottery tickets in a data-efficient way have been recently studied extensively as well. SNIP~\cite{lee2018snip} aims to find performant subnetworks with a few mini-batch iterations. GraSP~\cite{wang2020picking} and SynFlow~\cite{tanaka2020pruning} suggest that analyzing gradient-flow between layers enables identifying lottery tickets with a small set of training data or even without data.

Note that all of the pruning methods we discussed above mainly focus on 2D perception, i.e., image classification. There is a compression work on dense 3D ConvNets~\cite{Wang2020video}, but it is for video understanding to find spatiotemporal features not for spatial 3D data.
Hence, to the best of our knowledge, we use network compression methods for spatially sparse 3D neural networks for the first time for 3D perception.

Among the various categories of pruning methods, we mainly use \emph{unstructured pruning} since it is the most effective in finding highly compressed networks that have order-of-magnitude smaller parameters, which is a critical factor for 3D convnet compression.
Also, it can be applied to generic network architectures due to its simplicity, is simpler to implement, and can provide insights into structural properties from the final pruning patterns. 
\section{Preliminary}
\label{sec:preliminary}
Perception for spatial 3D data such as LiDAR point clouds or RGB-D scans requires non-standard operators that are not provided in the most off-the-shelf neural network libraries to handle sparsity and irregularity of 3D data and to implement fast and efficient local aggregation functions.
There are many representations and local aggregation functions, but in this work we focus on spatially sparse convnets. So, we cover the basics of a sparse tensor and spatially sparse convolution in this section.

\vspace{-4mm}
\subsubsection{Sparse Tensor} is the most common data structure for high-dimensional sparse data representation. It assumes that the majority of the data is 0 and saves only the non-zero elements by representing them as their coordinates and values. For high-dimensional tensors, the coordinate format (COO) is the most easy-to-use and versatile representation and we follow the convention.
Formally, a $D+1$ dimensional spatial sparse tensor consists of $D$ spatial dimensions and one dense feature dimension. 
Such sparse tensor with $N$ non-zero elements can be expressed as follows:
\begin{equation}
\label{eq:sparse_tensor}
\mathcal{T} = (\mathbf{C}, \mathbf{F}),
\end{equation}
where $\mathbf{C} \in \mathbb{Z}^{N \times D}$ is row-wise concatenation of discretized coordinates, and $\mathbf{F} \in \mathbb{R}^{N \times N_F}$ is corresponding feature vectors of size $N_F$. We will use $\mathbf{C}$ to denote a set of non-zero coordinates and a coordinate matrix, but it will be clear from the context.

\vspace{-4mm}
\subsubsection{Spatially Sparse Convolution} is a generalized version of the conventional dense convolution so that it can operate on a sparse tensor.
Given an input sparse tensor at $l$-th layer, $\mathcal{T}^l = (\mathbf{C}^l, \mathbf{F}^l)$ where $\mathbf{C}^l \in \mathbb{Z}^{N \times D}$ denotes the $D$-dimensional coordinates, and $\mathbf{F}^{l} \in \mathbb{R}^{N \times N_F^{l}}$ denotes corresponding features. The spatially spare convolution is defined as follows:
\begin{equation}
\label{eq:sparse_conv}
\mathbf{f}_{\mathbf{c}}^{l+1} = \sum_{\mathbf{i} \in \mathcal{K}(\mathbf{c})}\mathbf{W}_{\mathbf{i}}^{l} \cdot \mathbf{f}_{\mathbf{c+i}}^{l},
\end{equation}
where $\mathbf{f}^l_\mathbf{c} \in \mathbf{F}^l$ is a feature vector at coordinate $\mathbf{c}$, $\mathbf{W} \in \mathbb{R}^{K^D \times N_F^{l+1} \times N_F^{l}}$ is the kernel weights with kernel size $K$, $\mathbf{c} \in \mathbb{Z}^{D}$ is the current position of kernel center, and $\mathcal{K}(\mathbf{c})$ denotes a list of offsets in $D$-dimensional hypercube centered at $\mathbf{c}$.  
The convolution kernel, $\mathcal{K}(\cdot)$, is applied only on non-zero elements as the rest of the input is 0.

\vspace{-4mm}
\subsubsection{Neighbor Search}, where a convolution kernel finds non-zero neighbors within the current receptive field around $\mathbf{c}$ ($\mathcal{K}(\mathbf{c})$), is one of the areas the spatially sparse convolution shows its effectiveness and scalability. 
It uses a GPU hash table to speed up the search with $\mathcal{O}(1)$ complexity, which differentiates itself with other approaches that use continuous points and $\mathcal{O}(N^2)$ memory complexity for neighbor search~\cite{qi2017pointnet,qi2017pointnet++,thomas2019kpconv,zhao2021point,mao2019interpolated}.
Note that continuous convolution methods~\cite{thomas2019kpconv,xu2021paconv,zhao2021point} require multiple crops with a sliding window to handle large-scale scenes. Such sliding window makes them a few orders of magnitude slower than hash-table based neighbor search of spatially sparse convolution.

\vspace{-2mm}
\subsubsection{Spatial Sparsity vs. Weight Sparsity.}
The 3D convolutions are bulkier than 2D networks due to the extra spatial dimension. Note that in 2D, there are $K^2$ weight matrices of size $N_F^{l+1} \times N_F^{l}$ where $K$ is the kernel size, but in 3D, it requires $K^3$ weight matrices.
Thus, to reduce the number of parameters, we prune some of the weights from the convolution filters by applying successful pruning approaches~\cite{lecun1990optimal,lee2018snip,frankle2018the,frankle2019stabilizing,hassibi1993optimal,molchanov2016pruning,han2015learning}.
As a result, the pruned network has sparse weights, and we use them for convolution. To revisit \Eq{sparse_conv}, we prune the weight matrix $W_i$ and convert it to a sparse matrix. However, the convolution is applied only on spatially sparse coordinates $\mathbf{C}$. 

\section{Spatially Sparse Network Compression}
\label{sec:method}

Compressing a neural network requires first training a neural network with dense weights. We first discuss the target tasks and the training procedure in \Sec{3d_densenet}. Next, we introduce pruning methods we adopt in \Sec{pruning} and discuss the pruning procedure.
In \Sec{weight_sparse_convolution}, we use the compressed spatially sparse convnet or weight-sparse spatially sparse convnet (\OursAcronym) for faster inference by introducing our weight-sparse convolution algorithm.

\subsection{Training Spatially Sparse ConvNets}
\label{sec:3d_densenet}
\begin{figure*}[t!]
\centering
\includegraphics[width=0.95\textwidth]{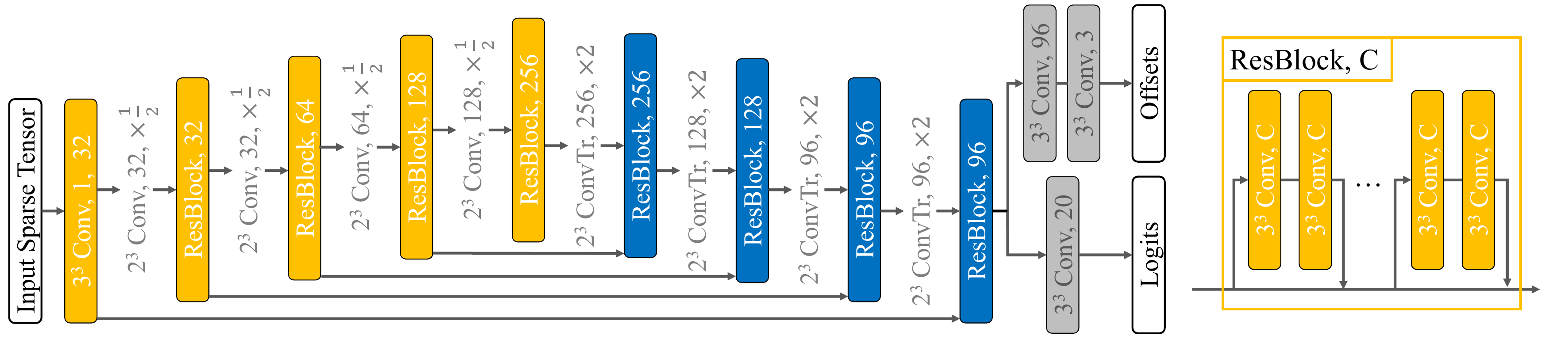}
\includegraphics[width=0.95\textwidth]{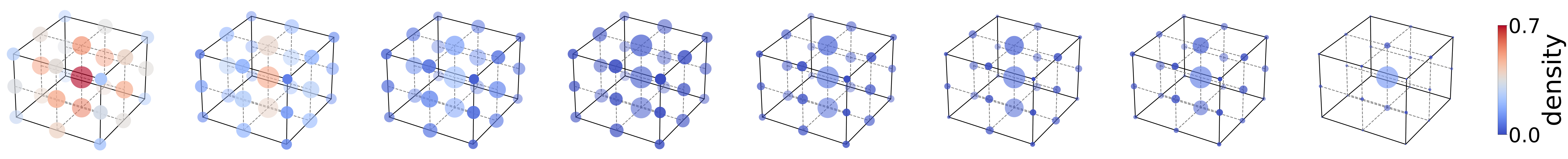}
\vspace{-2.0mm}
\caption{(Top) Res16UNet network architecture. It is a U-shape network with an encoder (yellow) and decoder (blue), and has skip connections.
(Bottom) Visualization of compressed convolution kernels. Each grid represents the average density of the first $3\times3\times3$ convolution kernel in each residual block from encoder to decoder from left to right. Note that kernels in decoder are pruned more aggressively in general, but weights along z-axis tend to survive}
\label{fig:kernel}
\end{figure*}

To achieve high performance and good network compression rate, neural network has to be fine-tuned on a specific task. We define two most common 3D tasks for our network compression that are practical and computationally demanding: semantic segmentation and instance segmentation.
For all experiments, we use U-shaped fully convolutional networks with residual blocks as depicted in \Fig{kernel} (top). For the last layer, we use logit prediction per voxel for semantic segmentation. For instance segmentation, we use logits with object center offset prediction as two outputs. We use a pytorch spatially sparse convolution implementation for all training~\cite{choy20194d}. 

We follow the standard training procedure in~\cite{choy20194d}, but do not apply any rotation averaging, or more advanced data augmentation techniques~\cite{nekrasov2021mix3d} to compare our network fairly with other methods that use the same datasets.
We use cross-entropy loss for semantic segmentation:
\begin{equation}
\label{eq:loss_semseg}
    \mathcal{L}_{\mathrm{sem}} = -\frac{1}{N}\sum_{n=1}^{N}\sum_{c=1}^{C}y_{n,c}\log({p_{n,c}}),
\end{equation}
where $N$ is the number of voxels, $C$ is the number of classes, $y_{n,c} = 1$ if the $n$th voxel belongs to $c$th class, otherwise 0, and $p_{n,c}$ indicates the predicted probability that $n$th voxel being class $c$.
For instance segmentation, we use the auxiliary loss for regressing the object center offset as follows:
\begin{equation}
\label{eq:loss_insseg}
    \mathcal{L}_{\mathrm{ins}} = \mathcal{L}_{\mathrm{sem}} 
    + \frac{1}{N}\sum_{n=1}^{N}\left( |\mathbf{d}_{n} -\mathbf{d}_{n}^{*}| -\frac{\mathbf{d}_n \cdot \mathbf{d}_{n}^*}{\|\mathbf{d}_n\| \|\mathbf{d}_{n}^{*}\|} \right)
\end{equation}
where $\mathbf{d}_{n}^{*}$ is the ground-truth displacement of $n$th voxel to the center of the instance that it belongs to, and $\mathbf{d}_{n}$ is the predicted displacement.
We train total three variants of networks with various depths and widths and train them with the same training procedure.
The detailed configuration of each network and the training parameters can be found in \Sec{implementation} and the supplementary material.

\begin{algorithm}[t]
\algsetup{linenosize=\small}
\small
    \DontPrintSemicolon
    \SetKwInOut{Input}{Input}
    \Input{$\mathcal{T}_{\mathrm{in}} = \{\mathbf{C}, \mathbf{F} \}$: an input sparse tensor with coordinates in COO layout. \\
    $\mathbf{C} \in \mathbb{Z}^{N \times 3}$ is non-zero discrete coordinates. \\
    $\mathbf{F} \in \mathbb{R}^{N \times N_{\mathrm{in}}}$ is a feature map of $\mathbf{C}$. \\
    $\mathbf{W} \in \mathbb{R}^{K \times N_{\mathrm{out}} \times N_{\mathrm{in}}}$: the weight tensor of sparse convolution layer with kernel size $K$, input feature dimension $N_{\mathrm{in}}$, output dimension $N_{\mathrm{out}}$ \\
    }
    \KwOut{$\mathcal{T}_{\mathrm{out}} = \{\mathbf{C}, \mathbf{F}'\}$: the output sparse tensor.}
    \SetKwBlock{Begin}{function}{end function}
    \Begin($\text{Weight-Sparse Convolution}$)
    {
        $\mathcal{K}(\cdot) \gets$ constructKernelMap($\mathbf{C}, K$) \tcp*[r]{construct kernel in-out map}
        
        $\mathbf{W}_{\mathrm{sp}}^T \gets\text{compress}(\text{transpose}(\mathbf{W}))$ \tcp*[r]{compress weights to CSR layout}
        
        $\mathbf{F}_{\mathrm{out}} \gets$ zeros$(shape=\{N, N_{\mathrm{out}}\})$ \tcp*[r]{initialize output features map}
        \For{$k \gets 1$ to $K^3$}
        {
            
            $(\mathcal{K}_{\mathrm{in}}, \mathcal{K}_{\mathrm{out}})\gets$ retrieveInOutMap$(\mathbf{C}, k)$ \tcp*[r]{retrieve kernel map}
            
            $\mathbf{F}_{\mathrm{in}}^T \gets \text{transpose}(\mathbf{F}[\mathcal{K}_{\mathrm{in}}, \colon])$ \tcp*[r]{transpose matrices}
            
            $\mathbf{F}_{\mathrm{out}}[\mathcal{K}_{\mathrm{out}}]$ += transpose$(\mathbf{W}_{\mathrm{sp},k}^T @ \mathbf{F}_{\mathrm{in}}^T)$ \tcp*[r]{sparse-dense matmul}
        }
        $\mathcal{T}_{\mathrm{out}} \gets \{\mathbf{C}, \mathbf{F}_{\mathrm{out}}\}$\\
        
        \Return{$\mathcal{T}_{\mathrm{out}}$}
    }\vspace{-1.0mm}
\caption{\textit{Weight-Sparse} Convolution Algorithm.}
\label{alg:weight_sparse_conv}
\end{algorithm}

\begin{wraptable}{R}{.50\columnwidth}
\centering
\small
\caption{Set of pruning strategies applied to the baseline models. We consider three different pruning criteria with \textit{Local} and \textit{Global} scope, which results in total six combinations. Here, $\mathcal{L}$ denotes the objective function, $\mathbf{w}_0$ is the initial, pretrained weights, and $\mathbf{w}_f$ denotes the weights that can be obtained after fine-tuning.
}
  \resizebox{.45\textwidth}{!}{
  \begin{tabular}{c|c||c|l}
    \toprule
    Tag & Approach & Pruning criteria & Scope \\ 
    \midrule
    L1L & L1 & $|\mathbf{w}_f|$ & Local \\
    L1G & L1 & $|\mathbf{w}_f|$ & Global \\
    FGL & First-order grad. & $|\frac{\partial \mathcal{L}}{\partial \mathbf{w}} \bigodot \mathbf{w}_f|$ & Local \\
    FGG & First-order grad. & $|\frac{\partial \mathcal{L}}{\partial \mathbf{w}} \bigodot \mathbf{w}_f|$ & Global \\
    L1SL & L1 w/ same sign & $\frac{\mathbf{w}_{0}}{|\mathbf{w}_{0}|} \cdot \mathbf{w}_{f}$ & Local \\
    L1SG & L1 w/ same sign & $\frac{\mathbf{w}_{0}}{|\mathbf{w}_{0}|} \cdot \mathbf{w}_{f}$ & Global \\
    \bottomrule
  \end{tabular}
  }
\label{tbl:pruning_method}
\end{wraptable}

\subsection{Network Pruning}
\label{sec:pruning}
Once we finish training a spatially sparse convnet with dense weights, we prune some of the weights while preserving the test accuracy.
In this paper, we utilize several pruning methods that belong to the family of magnitude-based pruning~\cite{lee2018snip,frankle2018the,zhou2019deconstructing,han2015learning}.
Specifically, we use three pruning methods with different pruning criteria and apply the criteria on the entire weights (global) or per layer (local). We put all our pruning criteria on \Tbl{pruning_method}.
Starting from a pre-trained model, we cycle through a series of the network pruning and fine-tuning iteratively.
At each pruning step, we use the pruning criteria to sort the weights and remove lower $p$ portion of weights and repeat this $N_\mathrm{prune}$ times so the final target pruning rate is $p_\mathrm{target} = 1 - (1 - p)^{N_\mathrm{prune}}$, where $p$ and $N_\mathrm{prune}$ are predefined hyperparameters.

We apply each pruning method to each pre-trained network with the same target compression rate to compare the effect of different pruning criteria. 
All hyperparameters other than pruning strategies, e.g., number of fine-tuning steps between consecutive pruning steps, learning rates, etc., are set to be the same for every model and a pruning strategy pair. 
We put the details of the pruning procedure on the supplementary material.

\subsection{Weight-Sparse Spatially Sparse Convolution}
\label{sec:weight_sparse_convolution}

Network compression of spatially sparse convnets results in highly compressed networks with minimal loss in accuracy. This leads to the reduction in memory footprint and faster transfer, but, faster inference requires dedicated software to make use of the compressed convolution kernels and the computational reduction from pruning.
To this end, we extend a pytorch spatially sparse convolution library~\cite{choy20194d} with a specialized convolution layer for weight-sparse spatially sparse convolution (WS${}^3$-conv). 
The WS${}^3$-conv consists of several steps:
\begin{itemize}[noitemsep,topsep=0pt,parsep=0pt,partopsep=0pt,leftmargin=*]
\renewcommand{\labelitemi}{$\bullet$}
\item \textit{Kernel mapping}: The spatially sparse convolution computes convolution only on non-zero elements neighbors. This requires extracting appropriate features from neighbors and placing the result on the center of convolution. This input neighbor to output pair is known as kernel mapping and we use it to select input features.
\item \textit{CSR compression}: We compress the dense weight matrices with pruning masks $m$. We use the \textit{Compressed Sparse Row} (CSR) format and convert the masked dense weight matrices to sparse matrices.
\item \textit{Transposition}: Majority of the off-the-shelf BLAS libraries such as \textit{Intel oneMKL} and \textit{cuSparse} commonly use sparse-dense matrix multiplication routine with the sparse matrix being the left operand. 
However, the feature maps are commonly represented in memory with row-major layout~\cite{choy20194d,pytorch}. 
Thus, we transpose the feature map for sparse dense multiplication. 
We pre-transpose weight matrices before the CSR compression. 
\item \textit{Sparse matrix multiplication}: Finally, we use off-the-shelf libraries to compute temporary features and place the output on the correct position defined by the kernel mapping.
\end{itemize}
We put the detailed algorithm on \Alg{weight_sparse_conv}.

\section{Experiment}
\label{sec:experiment}

\begin{figure}[!t]
\centering
\includegraphics[width=0.95\textwidth]{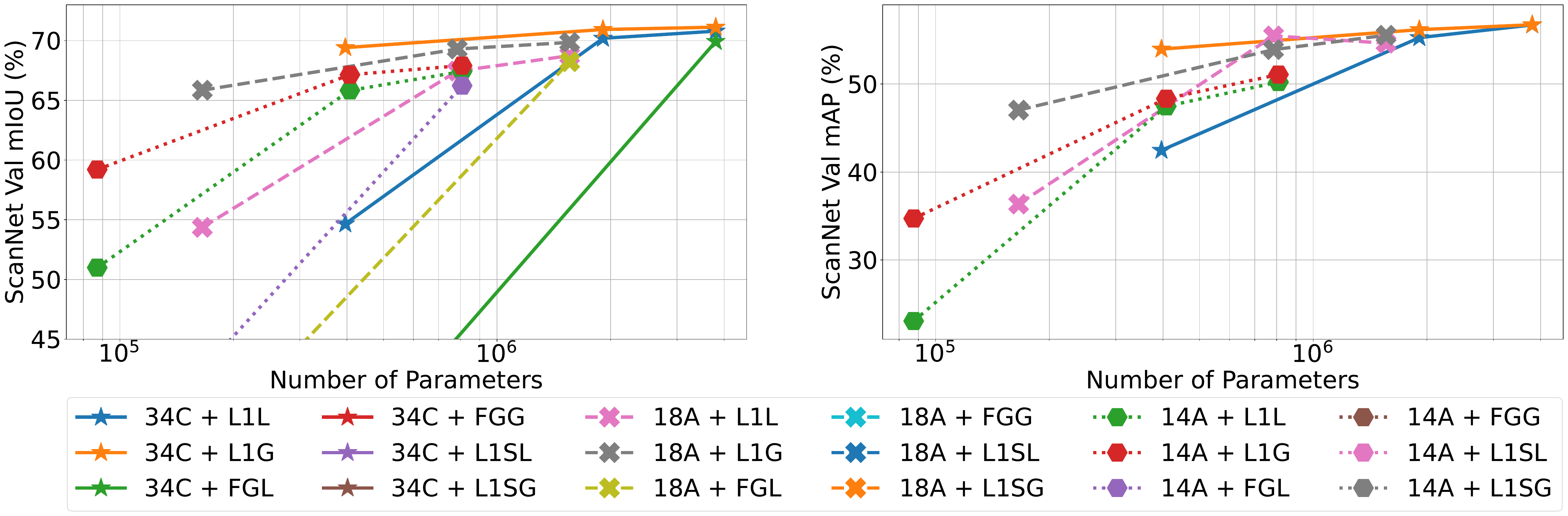}
\vspace{-2mm}
\caption{Ablation study on different pruning criteria on ScanNet semantic segmentation (Left) and instance segmentation (Right) tasks. For all architectures, L1G (Global L1 magnitude-based pruning) shows the least drop in performance. We use 34C, 18A, and 14A to denote Res16UNet variants. For pruning method tags, please refer to Table~\ref{tbl:pruning_method}}
\label{fig:ablation}
\end{figure}

\begin{table*}[t]
\setlength{\tabcolsep}{3pt}
\centering
\small
\caption{Quantitative results of \emph{semantic segmentation} models on ScanNet~\cite{dai2017scannet} and S3DIS dataset~\cite{armeni_cvpr16}. Gray row indicates the reference network, and we color green for the \textcolor{teal}{positive relative changes} and red for the \textcolor{purple}{negative relative changes} w.r.t. the reference network}
\vspace{-2mm}
  \resizebox{.95\textwidth}{!}{
  \begin{tabular}{l|c|c|c||c|c}  
        & & \multicolumn{2}{c||}{ScanNet~\cite{dai2017scannet}} & \multicolumn{2}{c}{S3DIS~\cite{armeni_cvpr16}} \\
        & \# Param.(M) & mIoU(\%) & mAcc(\%) & mIoU(\%) & mAcc(\%) \\ 
    \midrule
    PointNet~\cite{qi2017pointnet} & 3.5 (\nminus{56.6}) & 12.2 (\minus{82.2}) & 17.9 (\minus{77.0}) & 41.1 (\minus{37.8}) & 49.0 (\minus{34.6}) \\ 
    PointNet++~\cite{qi2017pointnet++}  & 2.0 (\nminus{75.2}) & 53.5 (\minus{21.9}) & - & 57.3 (\minus{13.3}) & 63.5 (\minus{15.3}) \\
    TangentConv~\cite{tatarchenko2018tangent} & 1.0 (\nminus{87.4}) & 40.9 (\minus{40.3}) & 55.1 (\minus{29.1}) & 52.8 (\minus{20.1}) & 62.2 (\minus{17.0}) \\
    PointConv~\cite{wu2019pointconv} & - & 61.0 (\minus{11.0}) & - & - & - \\
    PointASNL~\cite{yan2020pointasnl} & - & 63.5 (\minus{7.31}) & - & - & - \\
    KPConv~\cite{thomas2019kpconv} & 14.9 (\nplus{85.9}) & 69.2 (\plus{1.01}) & - & 67.1 (\plus{1.53}) & 72.8 (\minus{2.87}) \\
    PAConv~\cite{xu2021paconv}& 11.8 (\nplus{47.2}) & - & - & 66.6 (\plus{0.77}) & 73.0 (\minus{2.60})\\
    PointTransformer~\cite{zhao2021point} & 7.8 (\nminus{2.67}) & - & - & 70.4 (\plus{6.52}) & 76.5 (\plus{2.07})\\
    \midrule
    \rowcolor{gray!20}
    Res16UNet14A & 8.0 ($\pm$0.00) & 68.5 ($\pm$0.00) & 77.7 ($\pm$0.00) & 66.1 ($\pm$0.00) & 75.0 ($\pm$0.00)\\
    $\vdash$ 90\% pruning rate & 0.8 (\nminus{89.9}) & 67.9 (\minus{0.95}) & 77.3 (\minus{0.53}) & 65.1 (\minus{1.48}) & 72.0 (\minus{3.90}) \\
    $\vdash$ 95\% pruning rate & 0.4 (\nminus{94.9}) & 67.2 (\minus{1.96}) & 76.9 (\minus{1.08}) & 64.8 (\minus{1.91}) & 72.2 (\minus{3.64}) \\
    $\vdash$ 99\% pruning rate & 0.1 (\nminus{98.9}) & 59.3 (\minus{13.4}) & 70.3 (\minus{9.46}) & 60.9 (\minus{7.87}) & 68.7 (\minus{8.35})\\
    \midrule
    Res16UNet18A & 15.5 (\nplus{93.4}) & 70.3 (\plus{2.58}) & 79.7 (\plus{2.54}) & 67.6 (\plus{2.30}) & 74.6 (\minus{0.51}) \\
    $\vdash$ 90\% pruning rate & 1.6 (\nminus{80.6}) & 69.9 (\plus{2.03}) & 79.0 (\plus{1.72}) & 66.5 (\plus{0.57}) & 73.0 (\minus{2.60}) \\
    $\vdash$ 95\% pruning rate & 0.8 (\nminus{90.2}) & 69.3 (\plus{1.09}) & 78.6 (\plus{1.20}) & 66.3 (\plus{0.30}) & 73.2 (\minus{2.32}) \\
    $\vdash$ 99\% pruning rate & 0.2 (\nminus{97.9}) & 65.7 (\minus{4.10}) & 76.1 (\minus{2.03}) & 64.2 (\minus{2.84}) & 71.4 (\minus{4.68})\\
    \midrule 
    Res16UNet34C & 37.9 (\nplus{372.9}) & 71.6 (\plus{4.47}) & 80.4 (\plus{3.53}) & 68.8 (\plus{4.04}) & 75.8 (\plus{1.17}) \\
    $\vdash$ 90\% pruning rate & 3.8 (\nminus{52.6}) & 71.0 (\plus{3.68}) & 79.8 (\plus{2.70}) & 66.5 (\plus{0.59}) & 73.1 (\minus{2.44})  \\
    $\vdash$ 95\% pruning rate & 1.9 (\nminus{76.2}) & 71.0 (\plus{3.56}) & 79.7 (\plus{2.63}) & 66.4 (\plus{0.48}) & 73.2 (\minus{2.31}) \\
    $\vdash$ 99\% pruning rate & 0.4 (\nminus{95.1}) & 69.5 (\plus{1.40}) & 78.9 (\plus{1.60}) & 65.7 (\minus{0.67}) & 72.8 (\minus{2.92}) \\
  \end{tabular}
  }
\label{tbl:semseg}
\end{table*}

\begin{table*}[t]
\setlength{\tabcolsep}{3pt}
\centering
\small
\caption{Quantitative results of \emph{instance segmentation} models on ScanNet~\cite{dai2017scannet} and S3DIS dataset~\cite{armeni_cvpr16}. Gray row indicates the reference network, and we color green for the \textcolor{teal}{positive relative changes} and red for the \textcolor{purple}{negative relative changes} w.r.t. the reference network}
\vspace{-2mm}
  \resizebox{.95\textwidth}{!}{
  \begin{tabular}{l|c|c|c||c|c}  
        & & \multicolumn{2}{c||}{ScanNet~\cite{dai2017scannet}} & \multicolumn{2}{c}{S3DIS~\cite{armeni_cvpr16}} \\
        & \# Param.(M) & mAP\@50(\%) & mAP\@25(\%) & mAP\@50(\%) & mAP\@25(\%) \\ 
    \midrule
    \rowcolor{gray!20}
    Res16UNet14A & 8.0 ($\pm$0.00) & 52.8 ($\pm$0.00) & 68.1 ($\pm$0.00) & 54.5 ($\pm$0.00) & 61.7 ($\pm$0.00) \\
    $\vdash$ 90\% pruning rate & 0.8 (\nminus{89.9}) & 51.2 (\minus{3.03}) & 66.4 (\minus{2.50}) & 48.7 (\minus{10.6}) & 60.7 (\minus{1.62}) \\
    $\vdash$ 95\% pruning rate & 0.4 (\nminus{94.9}) & 48.9 (\minus{7.39}) & 65.4 (\minus{3.96}) & 48.7 (\minus{10.6}) & 59.4 (\minus{3.73}) \\
    $\vdash$ 99\% pruning rate & 0.1 (\nminus{98.9}) & 34.6 (\minus{34.5}) & 55.9 (\minus{17.9}) & 37.7 (\minus{30.8}) & 52.2 (\minus{15.4}) \\
    \midrule
    Res16UNet18A & 15.5 (\nplus{93.3}) & 55.8 (\plus{5.68}) & 71.3 (\plus{4.70}) & 55.6 (\plus{2.02}) & 62.7 (\plus{1.62}) \\
    $\vdash$ 90\% pruning rate & 1.6 (\nminus{80.5}) & 55.7 (\plus{5.49}) & 71.8 (\plus{5.43}) & 48.5 (\minus{11.0}) & 59.6 (\minus{3.40}) \\
    $\vdash$ 95\% pruning rate & 0.8 (\nminus{90.2}) & 55.5 (\plus{5.11}) & 71.2 (\plus{4.55}) & 49.8 (\minus{8.62}) & 58.7 (\minus{4.86}) \\
    $\vdash$ 99\% pruning rate & 0.2 (\nminus{97.9}) & 47.3 (\minus{10.4}) & 65.3 (\minus{4.11}) & 45.1 (\minus{17.2}) & 56.6 (\minus{8.27}) \\
    \midrule
    Res16UNet34C & 37.9 (\nplus{372.6}) & 57.0 (\plus{7.95}) & 73.1 (\plus{7.34}) & 54.7 (\plus{0.37}) & 64.1 (\plus{3.89}) \\
    $\vdash$ 90\% pruning rate & 3.8 (\nminus{52.6}) & 56.9 (\plus{7.77}) & 71.7 (\plus{5.29}) & 48.3 (\minus{11.4}) & 59.4 (\minus{3.73}) \\
    $\vdash$ 95\% pruning rate & 1.9 (\nminus{76.2}) & 56.3 (\plus{6.63}) & 71.5 (\plus{4.99}) & 51.0 (\minus{6.42}) & 60.0 (\minus{2.76}) \\
    $\vdash$ 99\% pruning rate & 0.4 (\nminus{95.0}) & 54.2 (\plus{2.65}) & 71.0 (\plus{4.26}) & 48.9 (\minus{10.3}) & 58.4 (\minus{5.35})\\
  \end{tabular}
  }
\label{tbl:insseg}
\end{table*}

\begin{table}[]
\centering
\small
\caption{Quantitative results of semantic segmentation models on SemanticKITTI dataset~\cite{behley2019iccv}. Gray row indicates the reference network, and we color green for the \textcolor{teal}{positive relative changes} and red for the \textcolor{purple}{negative relative changes} w.r.t. the reference network}
  \resizebox{.60\textwidth}{!}{
  \begin{tabular}{l|c|c|c}
     & \# Params.(M) & mIoU(\%) & mAcc(\%) \\ 
    \midrule
    \rowcolor{gray!20}
    Res16UNet34C & 37.9 ($\pm$0.00) & 61.6 ($\pm$0.00) & 68.5 ($\pm$0.00) \\
    $\vdash$ 90\% pruning rate & 3.8 (\nminus{90.0}) & 61.4 (\minus{0.32}) & 68.1 (\minus{0.58}) \\
    $\vdash$ 95\% pruning rate & 1.9 (\nminus{95.0}) & 60.4 (\minus{1.62}) & 67.2 (\minus{1.90})\\
    $\vdash$ 99\% pruning rate & 0.4 (\nminus{98.9}) & 56.1 (\minus{8.93}) & 64.4 (\minus{5.99})\\
    \midrule
    SPVCNN~\cite{tang2020searching} & 21.8 (\nminus{42.5}) & 63.7 (\plus{3.41}) & 70.7 (\plus{3.21}) \\
    $\vdash$ 90\% pruning rate & 2.2 (\nminus{94.2}) & 62.3 (\plus{1.14}) & 69.0 (\plus{0.73}) \\
    $\vdash$ 95\% pruning rate & 1.1 (\nminus{97.1}) & 61.4 (\minus{0.32}) & 67.3 (\minus{1.75}) \\
    $\vdash$ 99\% pruning rate & 0.2 (\nminus{99.5}) & 55.9 (\minus{9.25}) & 63.2 (\minus{7.73}) \\
    \midrule
    SPVNAS~\cite{tang2020searching} & 10.8 (\nminus{71.5}) & 64.7 (\plus{5.03}) & 72.2 (\plus{5.40}) \\
  \end{tabular}
  }
\label{tbl:semantickitti}
\end{table}

In this section, we present the results and analysis of our network pruning experiments for semantic and instance segmentation tasks on both real-world indoor and outdoor datasets.

\subsection{Experiment setup and metrics}
\label{sec:configuration}
We use semantic segmentation and instance segmentation on two indoor datasets and one outdoor dataset for the compression of spatially sparse convnets.
ScanNet~\cite{dai2017scannet} consists of $1.5$k scenes captured with commercial RGB-D cameras. It provides diverse indoor scenes but exhibits incomplete reconstruction and occlusion. On the other hand, the S3DIS~\cite{armeni_cvpr16} dataset provides 271 indoor office rooms captured with LiDAR 3D scanning devices. The reconstruction is complete and well aligned, but the scenes are monotonic compared with ScanNet.
Finally, SemanticKITTI~\cite{behley2019iccv} contains unparalleled number of scans that span the full 360 degree field-of-view of the employed automotive LiDAR with corresponding fine-grained semantic labels.

We measure the intersection of union (IoU) and per-point classification accuracy for each semantic class, and report the mean IoUs (mIoU), and mean accuracy (mAcc) for semantic segmentation.
For instance segmentation, we report average precision per class and mean average precision with varying accuracies 50\% and 25\% which we denote as mAP50 and mAP25 respectively.
To assess and compare the computational cost of pruned networks, we calculate the FLOPs and measure the per-scene latency on CPU.

We use three different network architectures with varying depth and width:
Res16UNet14A, Res16UNet18A, Res16UNet34C~\cite{choy20194d}. We apply unstructured pruning~\cite{han2015learning,lee2018snip,frankle2018the,zhou2019deconstructing} and compare the results with representative point-based approaches~\cite{qi2017pointnet,qi2017pointnet++,tatarchenko2018tangent,wu2019pointconv,yan2020pointasnl,thomas2019kpconv,xu2021paconv,zhao2021point}.
For outdoor semantic segmentation experiments, we evaluated another network architectures (SPVCNN \& SPVNAS)~\cite{tang2020searching}, hybrid architectures that equipped with both voxel-based and point-based branches, to validate that our findings are not strongly tied to the networks with specific architecture.

\subsection{Implementation detail}
\label{sec:implementation}
We train all our networks for $60K$ iterations with an initial learning rate $0.1$, $2cm$ voxelization, and batch size 8 on the indoor datasets.
After the baseline networks are converged, we apply our iterative pruning with  $i_{\mathrm{train}}=60K$ total training iteration, and $N_{\mathrm{prune}}=10$ pruning and fine-tuning sequences with $i_\mathrm{prune}=2K$ fine-tuning iterations between the consecutive pruning sequences for all experiments.
5cm voxelization and $i_\mathrm{train}=120K$ is used for outdoor semantic segmentation experiments.
For SPVCNN and SPVNAS~\cite{tan2019efficientnet}, we used the pretrained networks officially provided by the authors.
All the experiments are evaluated on an NVIDIA Titan RTX and Intel Xeon Gold 5220R.

\subsection{Experiment Results and Analysis}

We compress various backbone networks with multiple pruning rates and present the indoor semantic segmentation results on Table~\ref{tbl:semseg}, instance segmentation results on Table~\ref{tbl:insseg}, and outdoor semantic segmentation results on Table~\ref{tbl:semantickitti}. We present our analysis of these results in this section.

\vspace{-4mm}
\subsubsection{Conventional 3D ConvNets are bulky.}
On Table~\ref{tbl:semseg}, we can see that KPConv~\cite{thomas2019kpconv}, PAConv~\cite{xu2021paconv}, and PointTransformer~\cite{zhao2021point} achieve high performance on S3DIS while having as many parameters as Res16UNet14A. However, even when we prune 90\% of parameters from Res16UNet14A, the relative performance drop is only 0.5$\sim$3.9\%.
On the other hand, PointNet~\cite{qi2017pointnet}, PointNet++~\cite{qi2017pointnet++}, and TangentConv~\cite{tatarchenko2018tangent} have almost the same number of parameters as 90\% pruned Res16UNet14A. However, the 90\% pruned ResNet outperforms these networks by a decent margin. Similarly, PAConv~\cite{xu2021paconv} having 11.8M parameters shows comparable results to 90\% pruned Res16UNet18A with 1.6M parameters.

\vspace{-4mm}
\subsubsection{Prune overparametrized networks.}
Another interesting observation is that we see less performance drop when we compress heavier networks. For example, in Table~\ref{tbl:semseg}, Res16UNet14A with a 95\% compression (Network A) and Res16UNet34C with a 99\% compression (Network B) both have a similar number of parameters. Yet, Network B consistently shows higher accuracy on ScanNet and S3DIS.
We speculate that heavier and deeper networks still retain the same depths regardless of compression and overparametrization leads to a higher chance of maintaining more informative connections during pruning.

\vspace{-4mm}
\subsubsection{Global pruning is better than local pruning.}
We visualize the performance vs. network size graph on Figure~\ref{fig:ablation}. Global pruning that prunes the entire network weights perform better than local pruning that prunes the same percentage of weights from each layer. For instance, `34C+L1G' shows smaller drop in performance than `34C+L1L'. This is common in all backbone networks and we will explain this in the next paragraph.

\vspace{-4mm}
\subsubsection{Decoder of U-Net is overparametrized.}
We visualize the pruning patterns of convolution kernels on Figure~\ref{fig:kernel} (bottom). Note that the encoder retains more connections while the decoder is pruned aggressively. This indicates that the decoder has more redundant connections and shows that pruning globally allows the pruning algorithm to choose the importance freely among all layers leading to higher performance than local pruning.

\vspace{-4mm}
\subsubsection{Kernels along gravity-axis are more important.}
Pruning patterns on Figure~\ref{fig:kernel} (bottom) show that kernels along the gravity-axis have higher density than any other directions. We speculate that the vertical geometric patterns are especially important in indoor scenes due to structural similarity and height from the ground is critical in indoor perception.

\begin{figure*}[t!]
\centering
\resizebox{0.90\linewidth}{!}{
\includegraphics[width=\textwidth]{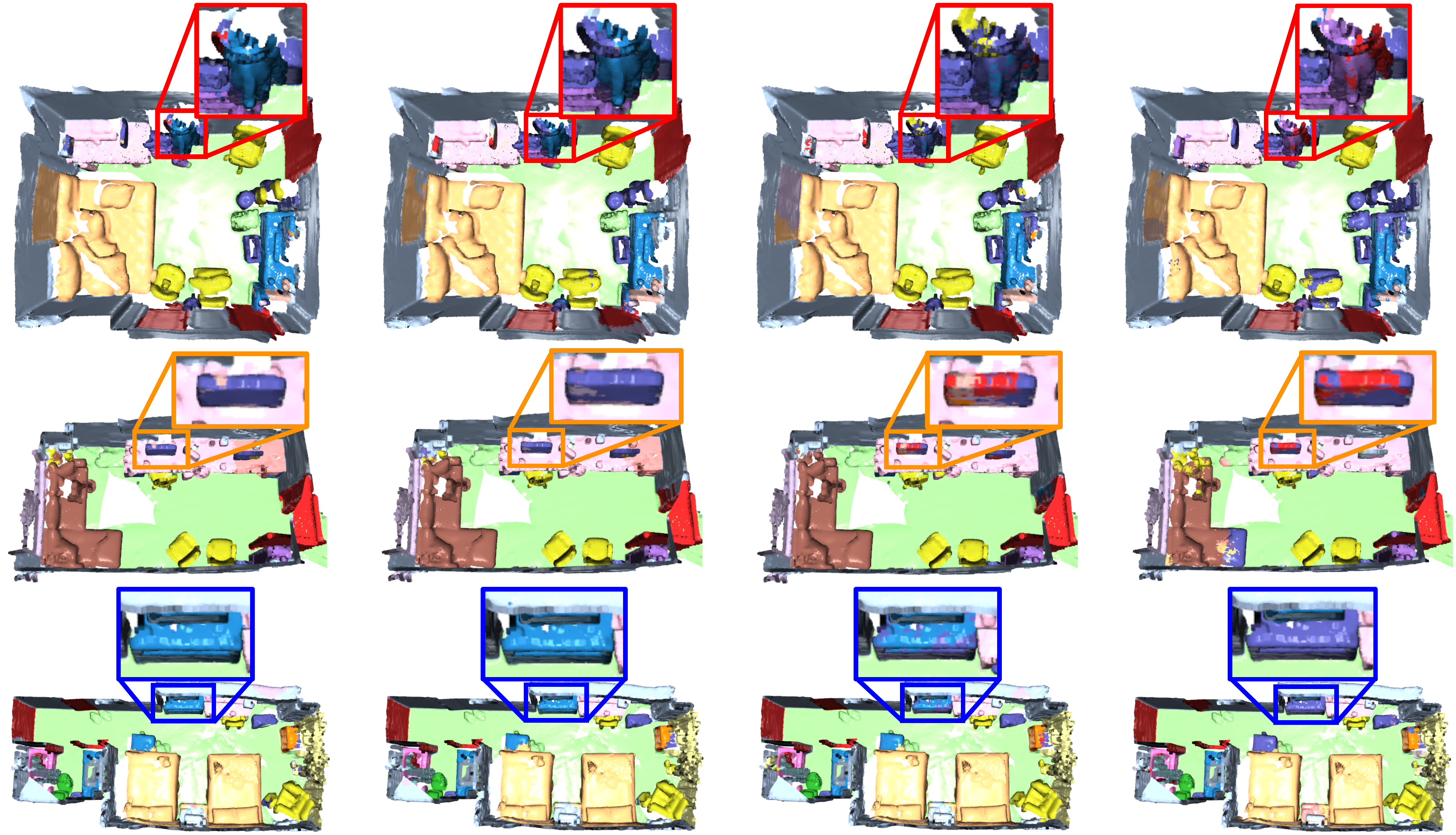}
}
\vspace{-2.0mm}
\caption{Visualization of semantic segmentation prediction on ScanNet~\cite{iandola2016squeezenet} using Res16UNet34C with varying sparsity (Dense, 90\%, 95\%, and 99\% pruned models from left to right).}
\label{fig:scannet_qual}
\end{figure*}

\subsection{\OursAcronym{}}
Based on the experiments and observations, we pick Res16UNet34C network with 99\% pruning rate as our representative result, and name it \OursAcronym{}. As shown in Table~\ref{tbl:semseg}, Table~\ref{tbl:insseg}, Figure~\ref{fig:teaser}, and \Fig{scannet_qual}, \OursAcronym{} shows very high accuracy (69.4\% mIoU) with only 0.4M parameters, which is even smaller than PointNet~\cite{qi2017pointnet} and PointNet++~\cite{qi2017pointnet++}. 

In addition, we test our weight-sparse convolution algorithm in Algorithm~\ref{alg:weight_sparse_conv} and measure the average FLOPs and per-scene latency on ScanNet~\cite{dai2017scannet} and SemanticKITTI~\cite{behley2019iccv} dataset and report the results on Table~\ref{tbl:flops}. 
Compared with the dense networks, the 99\% pruned \OursAcronym{} enables 95\% reduction in computational cost.
When the WS${}^3$-conv algorithm is enabled, it shows up to 30\% speedup by realizing the FLOPs reduction.

Furthermore, we show that the suggested pruning process successfully applies to another 3D convnet architecture (SPVCNN~\cite{tang2020searching}). This result indicates that our findings are not tied to a specific architecture.
We also put the results of  SPVNAS~\cite{tang2020searching}, an efficient variant of SPVCNN founded by network architecture search (NAS).
Despite its remarkable segmentation accuracy and moderate reduction in network parameters and FLOPs, the architecture search step is costly.
In general, it takes \emph{four times longer} than the standard training time. 
Specifically, if we need 15 epochs to train a baseline network, SPVNAS requires the following steps:
1) Training a super network for 15 epochs.
2) Taking additional 15 epochs to incorporate elastic depths.
3) Conducting evolutionary architecture search with numerous populations and generations.
4) Finetuning for about 10 to 15 epochs,
whereas our pruning procedure only requires moderate pruning and finetuning iterations when the pre-trained weights are accessible.

\subsection{Structured Pruning along Gravity-Axis}
Lastly, we utilize the fact that, on the decoder side, the convolution kernels along the gravity-axis are more important, and further compress the networks using structured pruning. We \emph{switch-off} the kernels that are off the z-axis and present the performance drop and speedup in \Tbl{ablation_zaxis}.
Note that switching off kernels in the last few layers doesn't harm the accuracy much, while it enables 45\% faster inference.

\begin{table}[]
\centering
\small
\vspace{-2mm}
\caption{Comparison of FLOPs, latency, and the number of parameters of semantic segmentation models trained and pruned on ScanNet~\cite{dai2017scannet} and SemanticKITTI~\cite{behley2019iccv}. Following~\cite{tang2020searching}, we calculate the average FLOPs per scene on the validation split for each dataset. As ~\cite{tang2020searching} stated, we do not consider operations other than convolution (e.g., BN) when calculating FLOPs. Latency is measured on \textbf{CPU}. 
}
  \resizebox{.98\textwidth}{!}{
  \begin{tabular}{l|c|c|r|r|r|r|r|r}
    & & & \multicolumn{3}{c|}{ScanNet~\cite{dai2017scannet}} & \multicolumn{3}{c}{SemanticKITTI~\cite{behley2019iccv}} \\
    \midrule
    Model & Prune & WS${}^3$-Conv & \# Params.(M) & \#FLOPs(G) & Latency(s) & \# Params.(M) & \#FLOPs(G) & Latency(s) \\ 
    \midrule
    \rowcolor{gray!20}
    Res16UNet34C & & & 37.9 ($\pm0.00$) & 201.0 ($\pm0.00$) & 4.7 ($\pm0.00$) & 37.9 ($\pm0.00$) & 294.8 ($\pm0.00$) & 6.3 ($\pm0.00$)\\
    $\vdash$ 99\% pruning rate & \checkmark & & 0.4 (\nminus{95.1}) & 9.6 (\nminus{95.2}) & 4.7 ($\pm0.00$) & 0.4 (\nminus{95.1}) & 14.8 (\nminus{95.0}) & 6.3 ($\pm0.00$) \\
    $\vdash$ 99\% pruning rate & \checkmark & \checkmark & 0.4 (\nminus{95.1}) & 9.6 (\nminus{95.2}) & 3.3 (\nminus{29.9})& 0.4 (\nminus{95.1}) & 14.8 (\nminus{95.0}) & 4.3 (\nminus{31.7}) \\
    \midrule
    SPVCNN & & & - & - & - & 21.8 (\nminus{42.5}) & 238.0 (\nminus{19.3}) & 7.2 (\nplus{14.3}) \\
    $\vdash$ 99\% pruning rate & \checkmark & & - & - & - & 0.2 (\nminus{99.5}) & 12.1 (\nminus{95.9}) & 7.2 (\nplus{14.3}) \\
    $\vdash$ 99\% pruning rate & \checkmark & \checkmark & - & - & - & 0.2 (\nminus{99.5}) & 12.1 (\nminus{95.9}) & 5.0 (\nminus{20.6}) \\
    \midrule 
    SPVNAS & & & - & - & - & 10.8 (\nminus{71.5}) & 130.0 (\nminus{55.9}) & $\ast$\footnotemark \\
    \midrule
  \end{tabular}
  }
\label{tbl:flops}
\end{table}

\begin{table}[t]
\setlength{\tabcolsep}{2pt}
\centering
\small
\vspace{-2mm}
\caption{Ablation study on the effect of structured pruning along z-axis aligned kernels. Columns B1 to B8 indicate residual blocks in our model (Res16UNet34C, 99\% pruned), $\checkmark$ denotes that the only weights along the z-axis of $3^3$ kernel are used to compute output for all conv layers in the block. Latency is measured on CPU. Gray row: reference dense (0\% pruned) model. Yellow row: the best network w/ 45\% faster inference. }
  \resizebox{.70\textwidth}{!}{
  \begin{tabular}{c|c|c|c|c|c|c|c|c|c|c}
     B1 & B2 & B3 & B4 & B5 & B6 & B7 & B8 & mIoU(\%) & mAcc(\%) & Latency(s) \\
    \midrule 
    \rowcolor{gray!20}
      &   &   &   &   &   &   &   & 71.6($\pm$0.00) & 80.4($\pm$0.00) & 4.71($\pm$0.00) \\ 
    \midrule
      &   &   &   &   &   &   &   & 69.5(\minus{2.93}) & 78.9(\minus{1.87}) & 3.30(\nminus{29.9}) \\
      \rowcolor{yellow!50}
      &   &   &   &   &   &   & \checkmark & 69.2(\minus{3.35}) & 78.8(\minus{1.99}) & 2.59(\nminus{45.0}) \\
      &   &   &   &   &   & \checkmark & \checkmark & 64.3(\minus{10.2}) & 73.5(\minus{8.58}) & 2.26(\nminus{52.0}) \\
      &   &   &   &   & \checkmark & \checkmark & \checkmark & 40.6(\minus{43.3}) & 46.0(\minus{42.8}) & 2.12(\nminus{55.0}) \\
      &   &   &   & \checkmark & \checkmark & \checkmark & \checkmark & 11.9(\minus{83.4}) & 15.7(\minus{80.5}) & 2.25(\nminus{52.2}) \\
      &   &   & \checkmark & \checkmark & \checkmark & \checkmark & \checkmark & 3.65(\minus{94.9}) & 7.10(\minus{91.2}) & 2.04(\nminus{56.7}) \\
      &   & \checkmark & \checkmark & \checkmark & \checkmark & \checkmark & \checkmark & 3.50(\minus{95.1}) & 6.92(\minus{91.4}) & 1.92(\nminus{59.2}) \\
      & \checkmark & \checkmark & \checkmark & \checkmark & \checkmark & \checkmark & \checkmark & 2.61(\minus{96.4}) & 6.03(\minus{92.5}) & 1.85(\nminus{60.2}) \\
    \checkmark & \checkmark & \checkmark & \checkmark & \checkmark & \checkmark & \checkmark & \checkmark & 5.63(\minus{92.1}) & 9.73(\minus{87.9}) & 1.72(\nminus{63.5}) \\
  \end{tabular}
  }
\label{tbl:ablation_zaxis}
\end{table}

\footnotetext{The layers in SPVNAS with dynamic depth and width do not support CPU inference}
\section{Discussion}
\label{sec:conclusion}
In this work, we apply various network compression methods on 3D spatially sparse convolutional networks. We found that 3D convnets can be compressed effectively to retain only 1\% of its original weights while maintaining very high accuracy.
Also, to speed up the inference, we propose weight-sparse convolution algorithm and show up to 45\% speedup when combined with structured pruning along gravity-axis.
Analysis shows interesting properties of 3D convnets and we further utilize the findings to apply additional structured pruning.

We only investigate the 3D segmentation networks, but would like to incorporate other 3D perception tasks, including object classification, object detection and registration tasks. Furthermore, based on our finding that there are inherent geometric patterns of significant weight parameters in \OursAcronym{}, we would like to investigate 3D perception-aware pruning techniques in future work. 

\clearpage
\bibliographystyle{unsrt}
\bibliography{egbib}

\clearpage
\section*{Supplementary Material}
This is a supplementary material for the paper, Putting 3D Spatially Sparse Networks on a Diet. We will further describe the details: architectural details (Sec. A), pruning algorithm (Sec. B), additional analysis on pruning criteria (Sec. C), additional kernel visualization (Sec. D), and additional quantitative and qualitative results (Sec.E).

\renewcommand{\theequation}{a.\arabic{equation}}
\renewcommand{\thetable}{a.\arabic{table}}
\renewcommand{\thefigure}{a.\arabic{figure}}
\renewcommand*{\thefootnote}{\arabic{footnote}}
\renewcommand\thesection{\Alph{section}}
\setcounter{section}{0}
\setcounter{figure}{0}
\setcounter{table}{0}

\section{Architecture details}
For all experiments, we used three baseline networks, namely Res16UNet14A, Res16UNet18A, and Res16UNet34C, for network compression and analysis. We provide the architectural details of the these networks. 
As illustrated in \Tbl{architecture}, the networks share the same backbone but have varying depths and widths.
The backbone has four steps of downsampling and upsampling using convolution and transposed convolution layers with stride 2.
Hence, at the very bottom of the bottleneck, the resolution of an input voxel reduces 16 times than its initial value.
In total, there are 8 residual blocks between upsampling and downsampling.
Between the same upsampled and downsampled features with the same resolution, it has skip connections.
For semantic segmentation, the features are projected to logits using $1\times1\times1$ convolution layer with $N_C$ output dimension, where $N_C = 20$ for ScanNet~\cite{dai2017scannet}, 13 for S3DIS~\cite{armeni_cvpr16}, and 19 for SemanticKITTI~\cite{behley2019iccv}.
Note that instance segmentation models have an additional head for predicting the coordinate offsets for instance centroids.

\section{Pruning Algorithm}
In Alg~\ref{alg:pruning}, we put the detailed algorithm of our iterative pruning and finetuning procedure. 
After performing an extensive hyperparameter searches, we found that $i_\mathrm{prune}=2K$, $N_\mathrm{prune}=10$ produce the optimal pruned networks with minimal drop in accuracy across all networks, tasks, and datasets, hence we fixed those parameters for our main experiments.
$i_\mathrm{train}$ and $p_\mathrm{target}$ are then determined for each experiment.
\begin{algorithm}[t]
\algsetup{linenosize=\small}
\small
    \DontPrintSemicolon
    \SetKwInOut{Input}{Input}
    \Input{$\mathcal{F}(\cdot,\theta_{\mathrm{init}})$: a pre-trained net. with param. $\theta_{\mathrm{init}}$. \\
    $i_{\mathrm{train}}$: number of training iterations. \\
    $i_\mathrm{prune}$: number of fine-tuning iterations per step.\\
    $N_{\mathrm{prune}}$: number of pruning iteration to repeat. \\
    $p_{\mathrm{target}}$: target pruning rate. \\
    criteria$(\cdot)$: a score function for pruning.\\
    }
    \KwOut{$\theta'$: a pruned network parameter.}
    \SetKwBlock{Begin}{function}{end function}
    \SetKwRepeat{Do}{do}{while}
    \Begin(IterativePruning)
    {
        $\theta \gets \theta_{\mathrm{init}}$\\
        $m \gets $ ones\_like($\theta$)~~\tcp{init. pruning mask} %
        $p \gets 1 - (1-p_{\mathrm{target}})^{\frac{1}{N_{\mathrm{prune}}}}$\\
        \For{$i \gets 1$ to $i_{\mathrm{train}}$}
        {
            \eIf{$i \modd i_{\mathrm{prune}} == 0$}
            {
                $s \gets $ criteria($\theta \bigodot m$)~~\tcp{assign score}
                $\mathrm{idx} \gets $ argsort($s,p$)~~\tcp{sort parameters by score}
                $m[\mathrm{idx}] \gets 0$~~\tcp{update pruning mask}
            }
            {
                $\theta \gets$ train($\mathcal{F}(\cdot, \theta \bigodot m)$)~~\tcp{fine-tune}
            }
        }
        \Return{$\theta' \gets \theta$}
    }
\caption{Iterative Pruning Algorithm.}
\label{alg:pruning}
\end{algorithm}

\section{Additional analysis of pruning criteria}

\begin{figure}[!htb]
\centering
\includegraphics[width=0.9\textwidth]{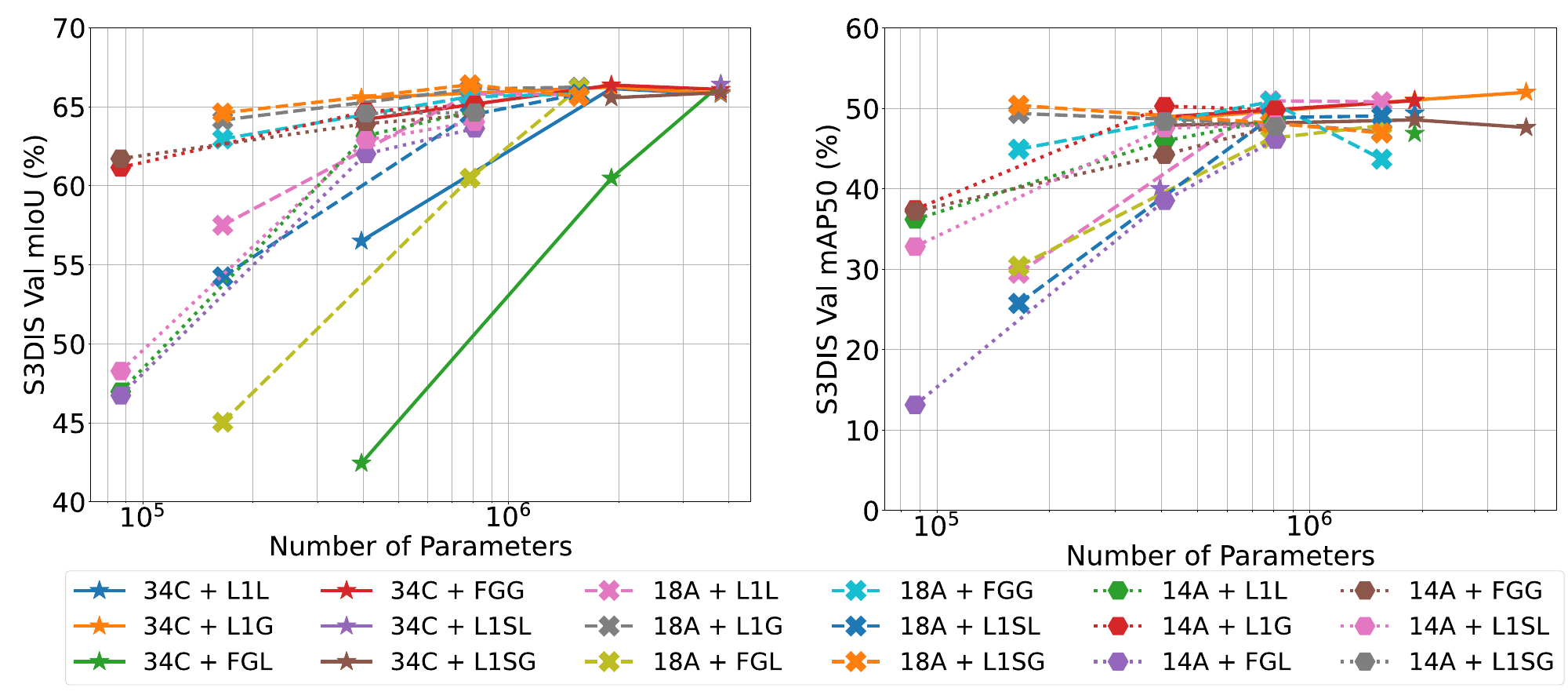}
\caption{Ablation study on different pruning criteria on S3DIS~\cite{armeni_cvpr16} semantic segmentation (Left), and S3DIS instance segmentation (Right).}
\label{fig:prune_strat}
\end{figure}

In \Fig{prune_strat}, we visualize the performance vs. model size graphs on S3DIS semantic segmentation (left), and S3DIS instance segmentation (right).
For semantic segmentation model, we use mIoU for our main performance metric; for instance segmentation, we use mAP50 for our evaluation metric.

We found that the global pruning strategies show better pruning efficiency than the local pruning methods across all datasets and tasks.
Note that in the S3DIS instance segmentation experiment, some pruning strategies show the best performance on a 95\% pruned model, not on 90\% pruned models.
We speculate that the 95\% pruned models generalize better than the larger 90\% pruned models because the instance segmentation task on S3DIS is monotonic and lack diversity compared with the ScanNet dataset.

\section{Additional kernel visualization}

\begin{figure*}[t]
\centering
\small
\resizebox{0.95\linewidth}{!}{
\begin{tabular}{c|c|c|c}
    Criteria & \begin{tabular}{@{}c@{}}Prune\\rate\end{tabular} & Kernel Visualization & \begin{tabular}{@{}c@{}}mIoU\\ (\%)\end{tabular} \\
    \toprule
    L1L
    & 99\% &
    \begin{tabular}{@{}c@{}}\includegraphics[width=0.9\textwidth]{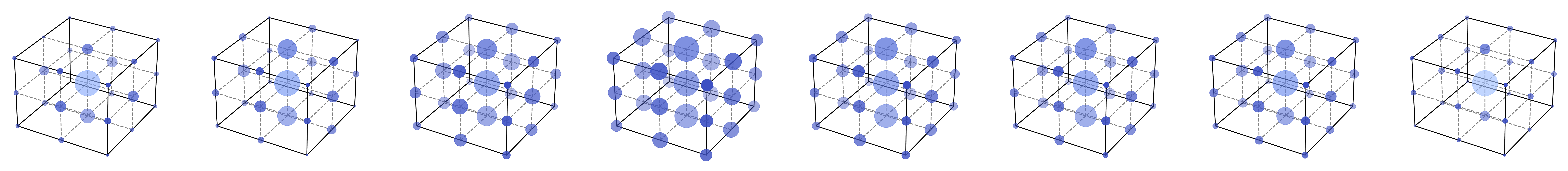}\end{tabular} & 54.7 \\
    
    FGL
    & 99\% &
    \begin{tabular}{@{}c@{}}\includegraphics[width=0.9\textwidth]{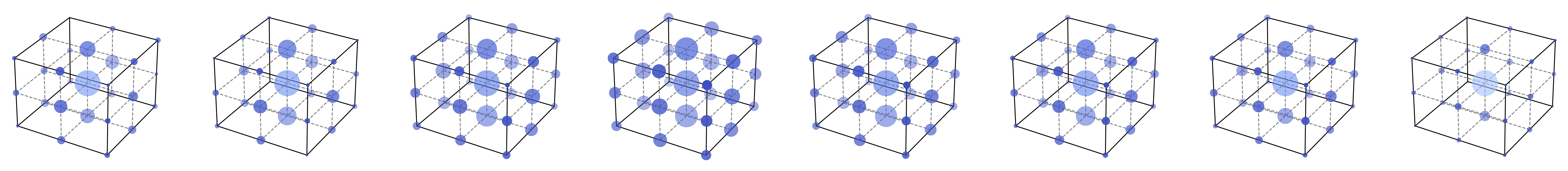}\end{tabular} & 34.4 \\
    
    FGG
    & 99\% & 
    \begin{tabular}{@{}c@{}}\includegraphics[width=0.9\textwidth]{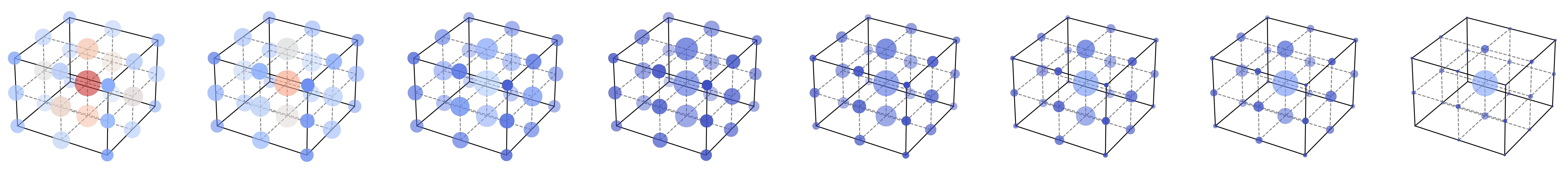}\end{tabular} & 68.5  \\
    
    L1SL
    & 99\% &
    \begin{tabular}{@{}c@{}}\includegraphics[width=0.9\textwidth]{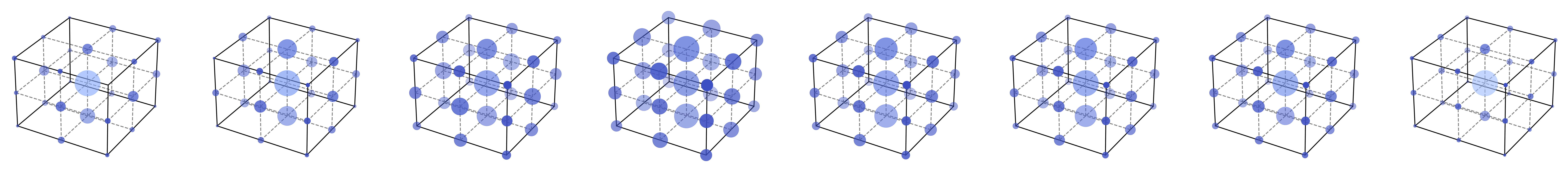}\end{tabular} & 58.1 \\

    L1SG
    & 99\% & 
    \begin{tabular}{@{}c@{}}\includegraphics[width=0.9\textwidth]{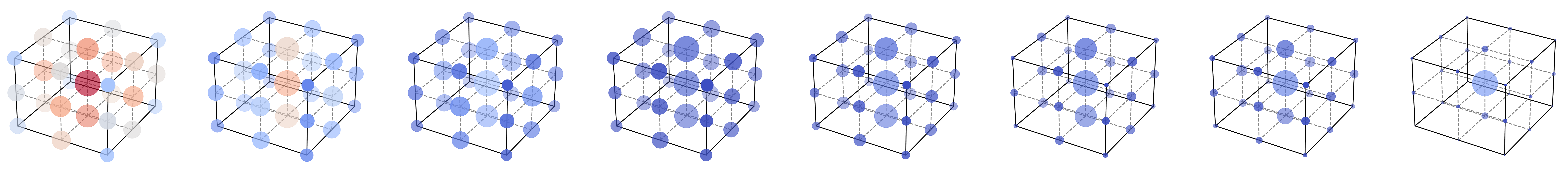}\end{tabular} & 68.9 \\
    
    L1G
    & 99\% & 
    \begin{tabular}{@{}c@{}}\includegraphics[width=0.9\textwidth]{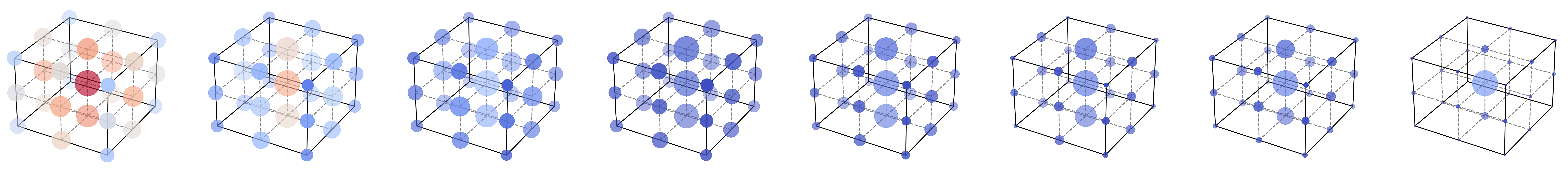}\end{tabular} & 69.5 \\
    \midrule
    L1G
    & 95\% & 
      \begin{tabular}{@{}c@{}}\includegraphics[width=0.9\textwidth]{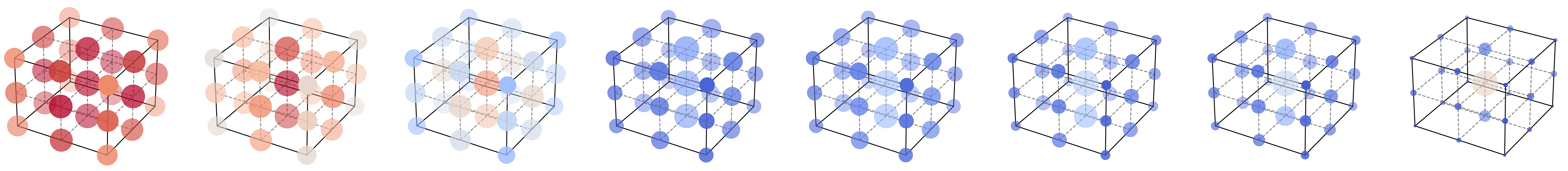}\end{tabular} & 71.0 \\
    L1G
    & 90\% & 
      \begin{tabular}{@{}c@{}}\includegraphics[width=0.9\textwidth]{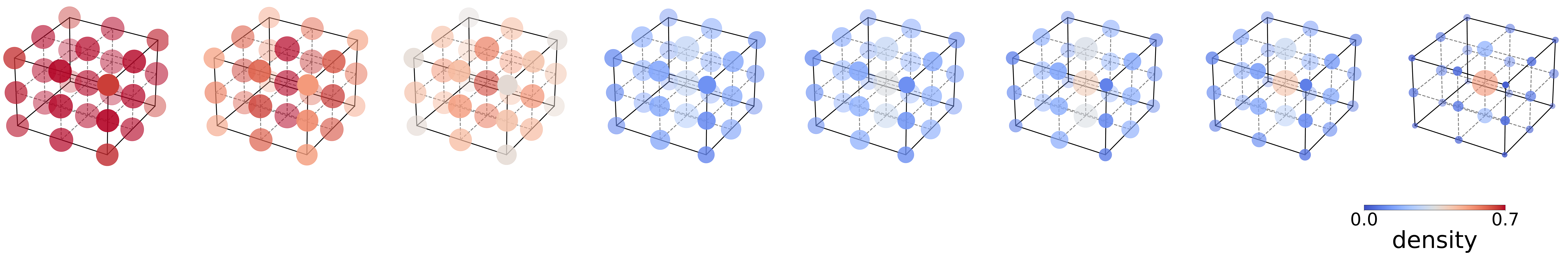}\end{tabular} & 71.1 \\
    \bottomrule
\end{tabular}
}
\caption{Visualizations of compressed convolution kernels. We use Res16UNet34C for all kernel visualizations. We present convolution kernels compressed with different pruning criteria on the first 6 rows, and kernels compressed with different compression rate on the last three rows.}
\label{fig:kernel_34c}
\end{figure*}

In \Fig{kernel_34c}, we visualize the compressed convolution kernels with different pruning strategies, or different compression rates. All models are Res16UNet34C trained and pruned on ScanNet semantic segmentation.

As shown in the figure, all pruning strategies tend to prune the weights of decoder more aggressively.
While the models pruned with global pruning methods have relatively dense encoder, the local pruning methods make the encoder sparser.
Furthermore, we identify that the pattern of convolution kernels, alignment along the gravity axis, reveals as the model is compressed more significantly.

\section{More quantitative and qualitative results}
\begin{figure*}[t]
\centering
\small
\resizebox{1.0\linewidth}{!}{
\begin{tabular}{cccc}
    \includegraphics[width=0.24\textwidth]{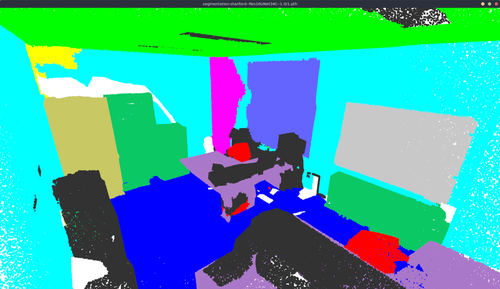} &
    \includegraphics[width=0.24\textwidth]{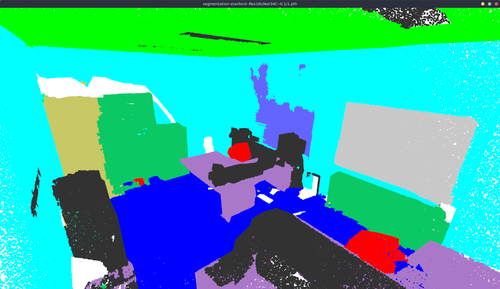} &
    \includegraphics[width=0.24\textwidth]{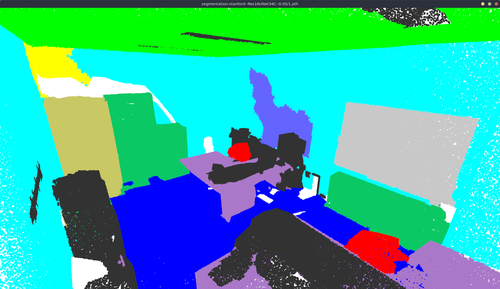} &
    \includegraphics[width=0.24\textwidth]{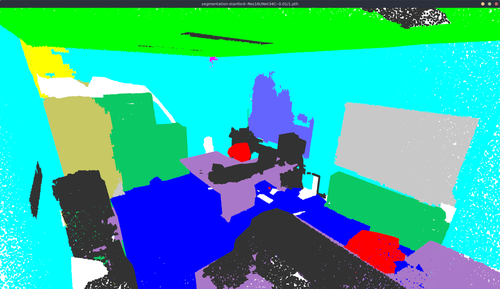} \\
    \includegraphics[width=0.24\textwidth]{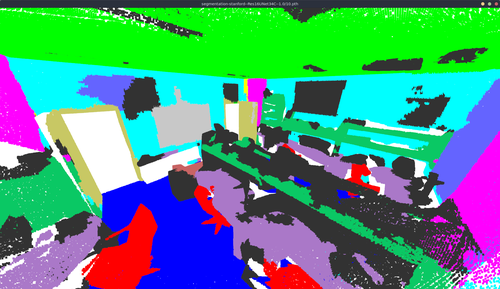} &
    \includegraphics[width=0.24\textwidth]{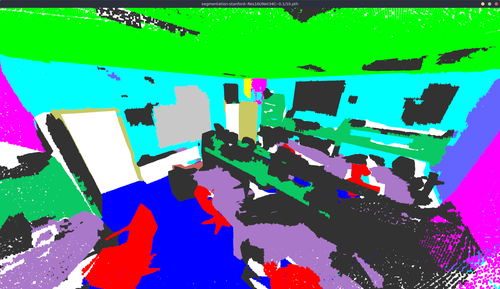} &
    \includegraphics[width=0.24\textwidth]{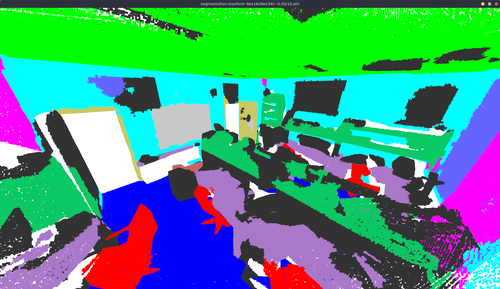} &
    \includegraphics[width=0.24\textwidth]{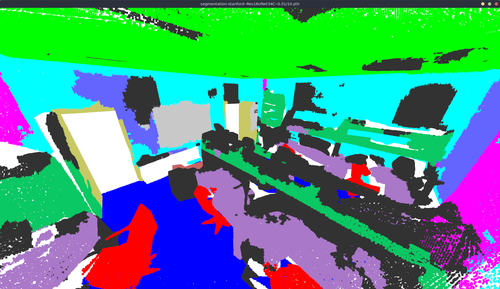} \\
    \includegraphics[width=0.24\textwidth]{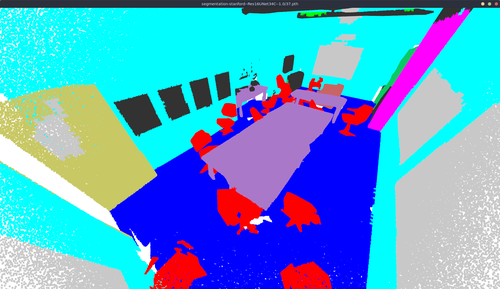} &
    \includegraphics[width=0.24\textwidth]{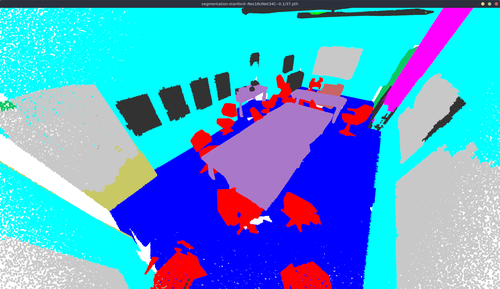} &
    \includegraphics[width=0.24\textwidth]{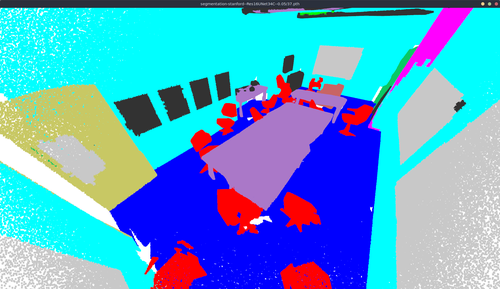} &
    \includegraphics[width=0.24\textwidth]{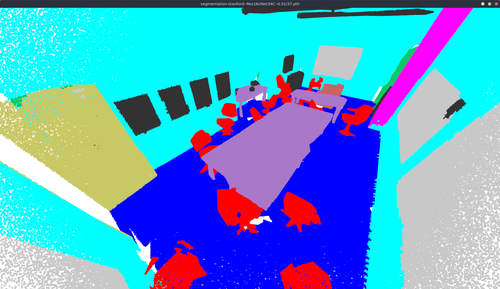} \\
    \begin{tabular}{@{}c@{}}(a) Reference Network \\ (Res16UNet34C) \end{tabular} & (b) 90\% pruned & (c) 95\% pruned & (d) 99\% pruned \\ 
\end{tabular}
}
\caption{Semantic segmentation qualitative results on S3DIS~\cite{armeni_cvpr16}.}
\label{fig:stanford_semseg}
\end{figure*}

\begin{figure*}[t]
\centering
\small
\resizebox{1.0\linewidth}{!}{
\begin{tabular}{cccc}
    \includegraphics[width=0.24\textwidth]{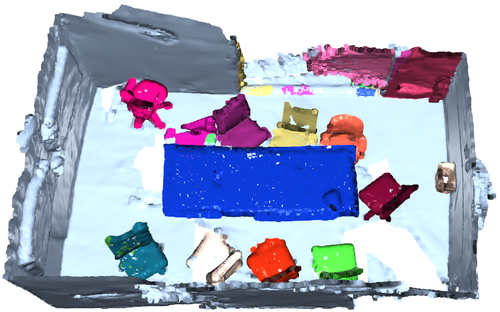} &
    \includegraphics[width=0.24\textwidth]{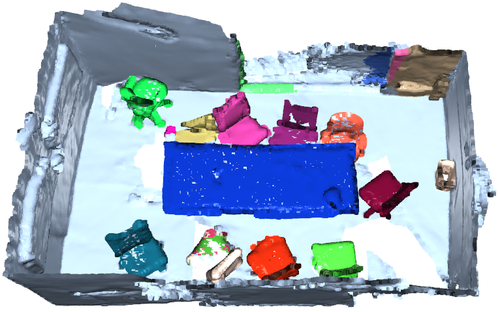} &
    \includegraphics[width=0.24\textwidth]{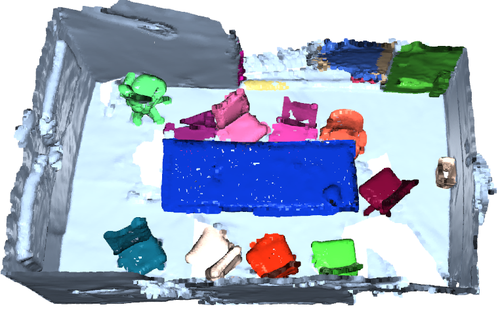} &
    \includegraphics[width=0.24\textwidth]{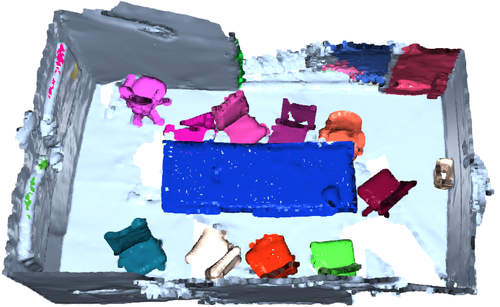} \\
    \includegraphics[width=0.24\textwidth]{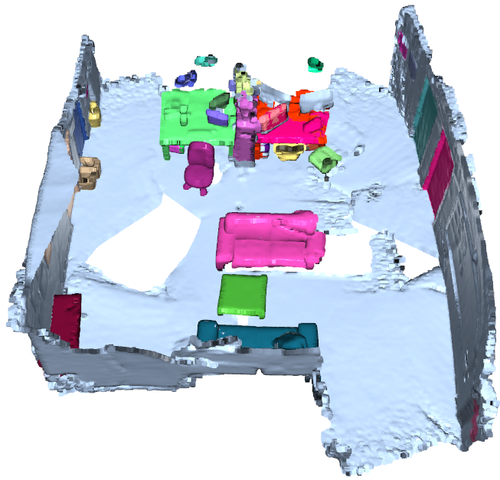} &
    \includegraphics[width=0.24\textwidth]{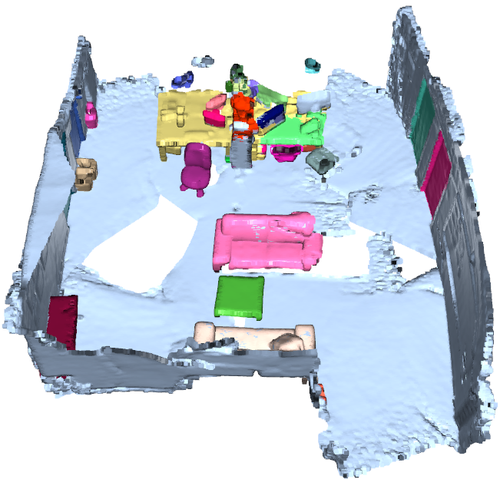} &
    \includegraphics[width=0.24\textwidth]{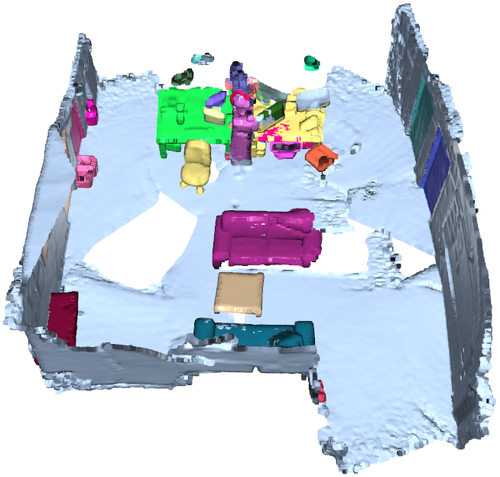} &
    \includegraphics[width=0.24\textwidth]{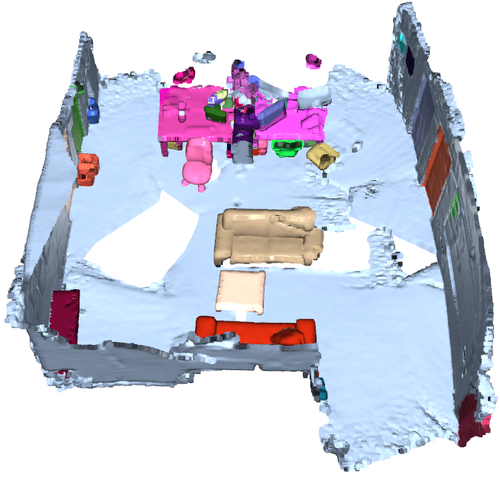} \\
    \includegraphics[width=0.24\textwidth]{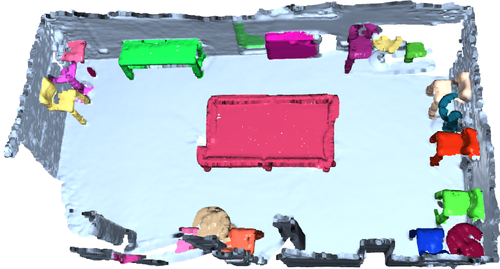} &
    \includegraphics[width=0.24\textwidth]{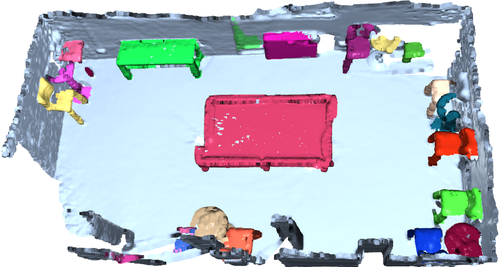} &
    \includegraphics[width=0.24\textwidth]{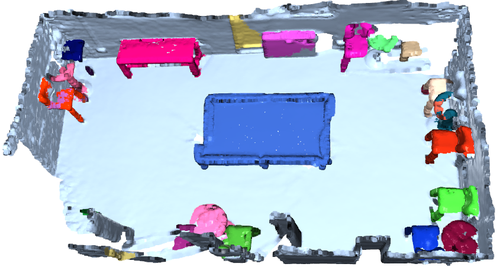} &
    \includegraphics[width=0.24\textwidth]{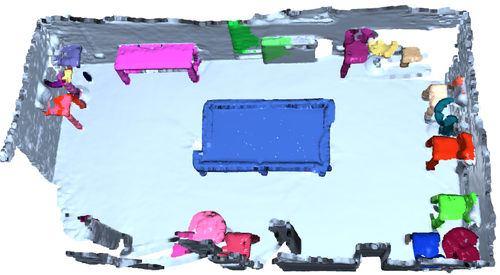} \\
    \begin{tabular}{@{}c@{}}(a) Reference Network \\ (Res16UNet34C) \end{tabular} & (b) 90\% pruned & (c) 95\% pruned & (d) 99\% pruned \\ 
\end{tabular}
}
\caption{Instance segmentation qualitative results on ScanNet~\cite{dai2017scannet}.}
\label{fig:scannet_insseg}
\end{figure*}

\begin{figure*}[t]
\centering
\small
\resizebox{1.0\linewidth}{!}{
\begin{tabular}{cccc}
    \includegraphics[width=0.24\textwidth]{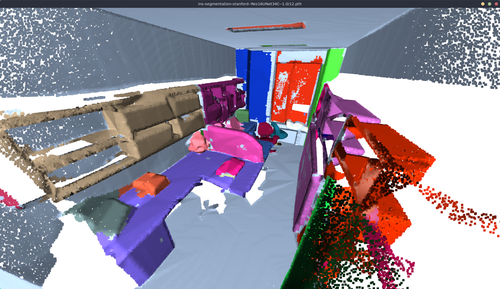} &
    \includegraphics[width=0.24\textwidth]{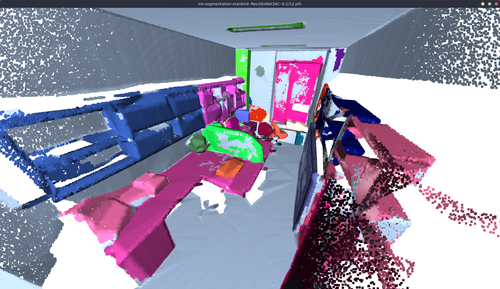} &
    \includegraphics[width=0.24\textwidth]{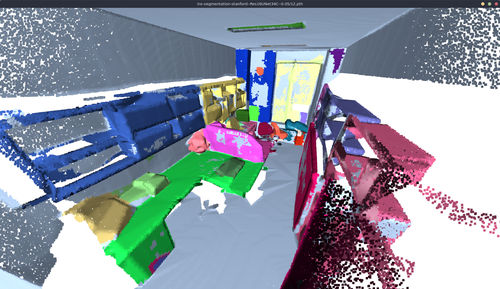} &
    \includegraphics[width=0.24\textwidth]{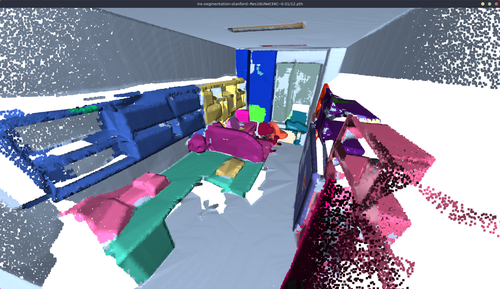} \\
    \includegraphics[width=0.24\textwidth]{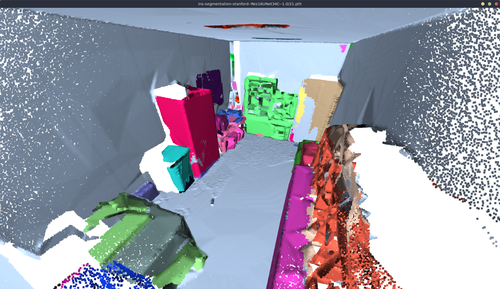} &
    \includegraphics[width=0.24\textwidth]{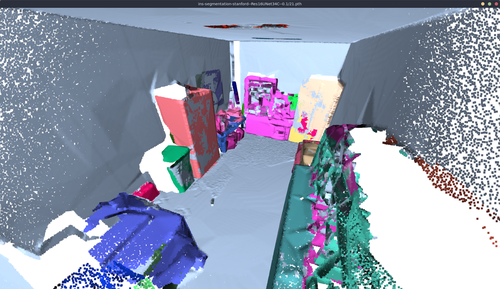} &
    \includegraphics[width=0.24\textwidth]{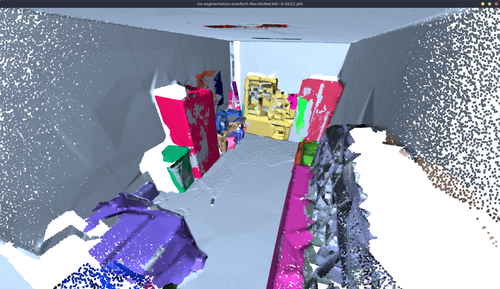} &
    \includegraphics[width=0.24\textwidth]{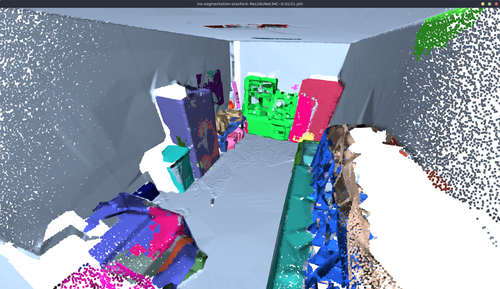} \\
    \includegraphics[width=0.24\textwidth]{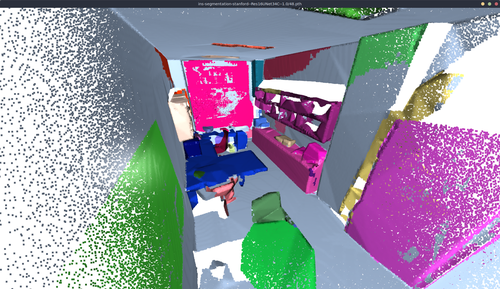} &
    \includegraphics[width=0.24\textwidth]{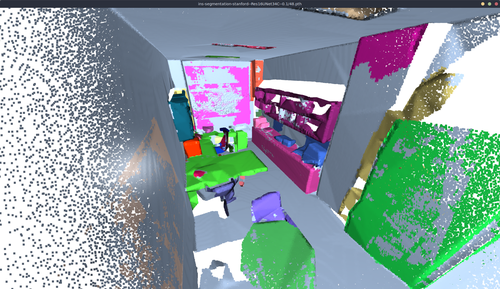} &
    \includegraphics[width=0.24\textwidth]{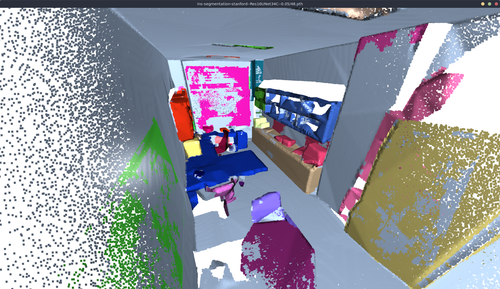} &
    \includegraphics[width=0.24\textwidth]{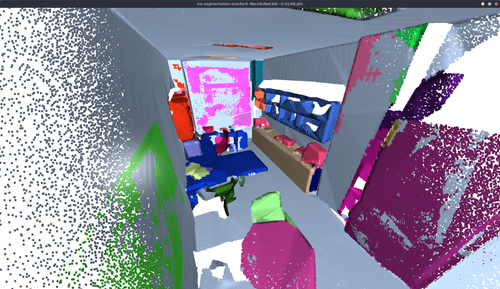} \\
    \includegraphics[width=0.24\textwidth]{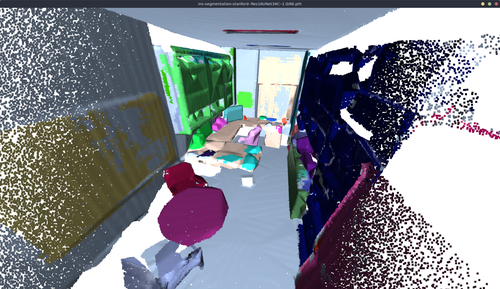} &
    \includegraphics[width=0.24\textwidth]{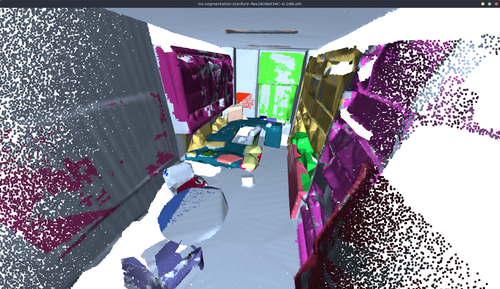} &
    \includegraphics[width=0.24\textwidth]{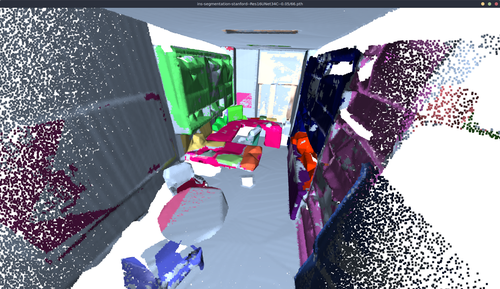} &
    \includegraphics[width=0.24\textwidth]{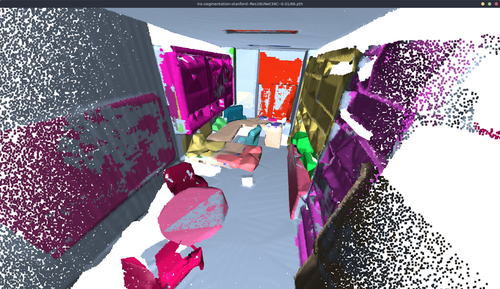} \\
    \begin{tabular}{@{}c@{}}(a) Reference Network \\ (Res16UNet34C) \end{tabular} & (b) 90\% pruned & (c) 95\% pruned & (d) 99\% pruned \\ 
\end{tabular}
}
\caption{Instance segmentation qualitative results on S3DIS~\cite{armeni_cvpr16}.}
\label{fig:stanford_insseg}
\end{figure*}

\begin{figure*}[t]
\centering
\small
\resizebox{1.0\linewidth}{!}{
\begin{tabular}{cccc}
    \includegraphics[width=0.24\textwidth]{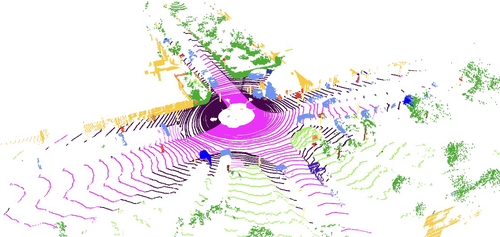} &
    \includegraphics[width=0.24\textwidth]{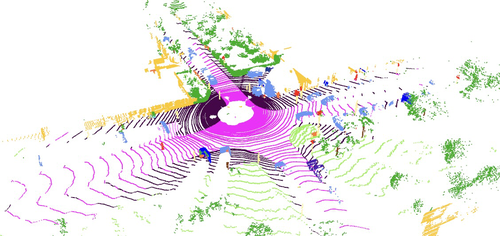} &
    \includegraphics[width=0.24\textwidth]{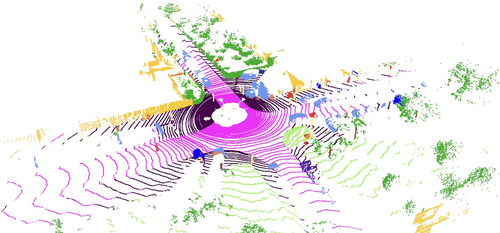} &
    \includegraphics[width=0.24\textwidth]{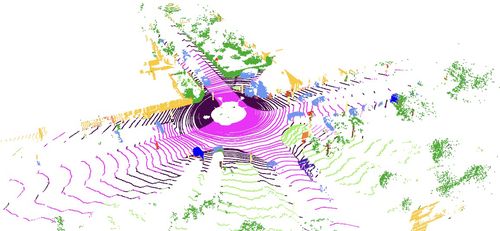} \\
    \includegraphics[width=0.24\textwidth]{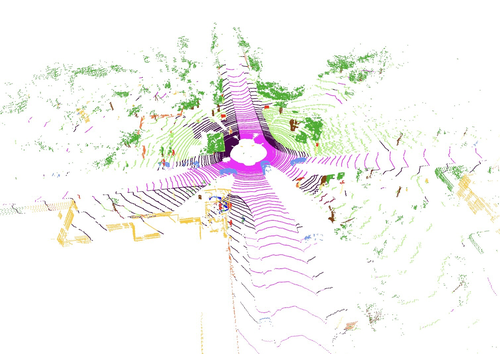} &
    \includegraphics[width=0.24\textwidth]{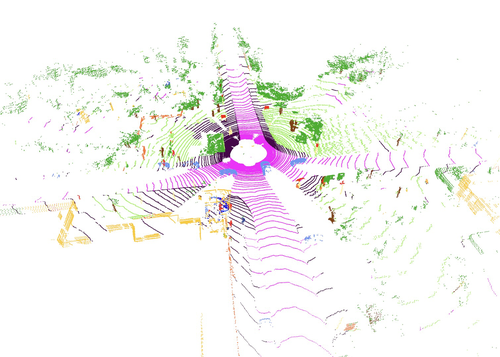} &
    \includegraphics[width=0.24\textwidth]{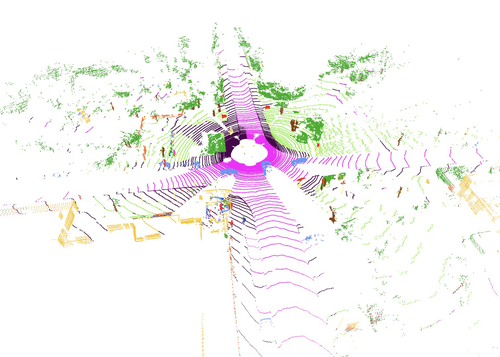} &
    \includegraphics[width=0.24\textwidth]{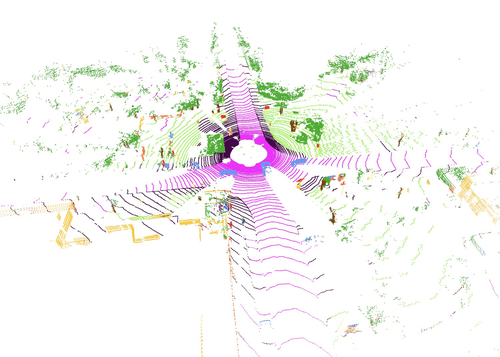} \\
    \includegraphics[width=0.24\textwidth]{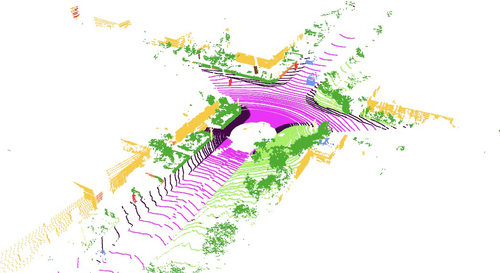} &
    \includegraphics[width=0.24\textwidth]{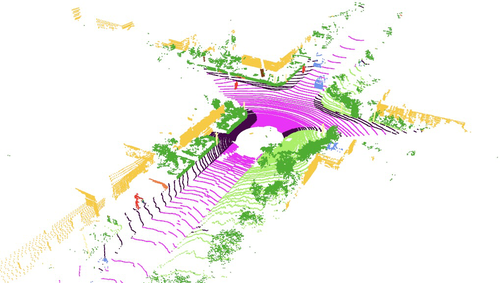} &
    \includegraphics[width=0.24\textwidth]{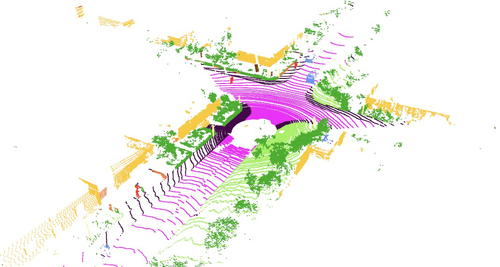} &
    \includegraphics[width=0.24\textwidth]{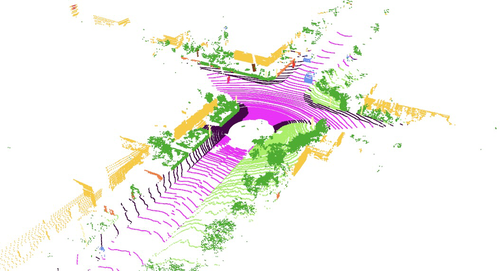} \\
    \includegraphics[width=0.24\textwidth]{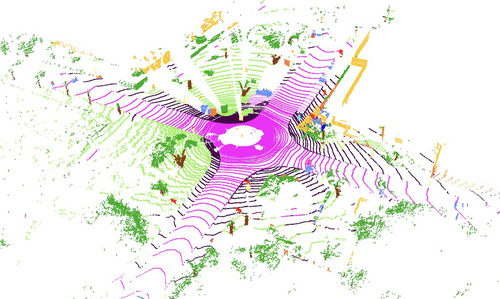} &
    \includegraphics[width=0.24\textwidth]{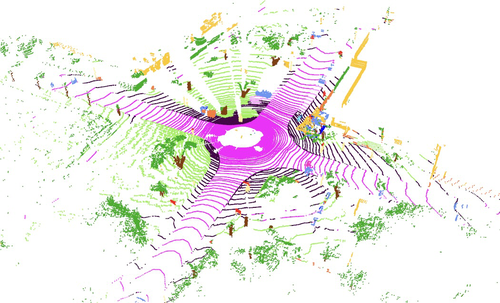} &
    \includegraphics[width=0.24\textwidth]{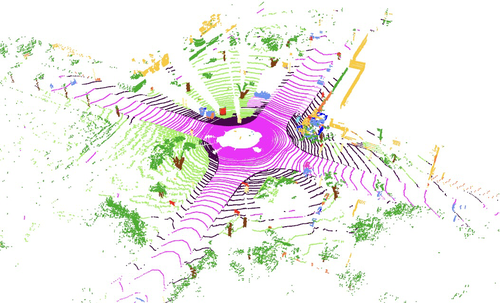} &
    \includegraphics[width=0.24\textwidth]{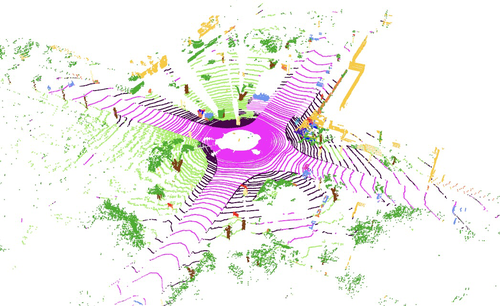} \\
    \begin{tabular}{@{}c@{}}(a) Reference Network \\ (Res16UNet34C) \end{tabular} & (b) 90\% pruned & (c) 95\% pruned & (d) 99\% pruned \\ 
\end{tabular}
}
\caption{Semantic segmentation qualitative results on SemanticKITTI~\cite{behley2019iccv}.}
\label{fig:kitti_semseg}
\end{figure*}

We provide additional scenewise semantic segmentation results on ScanNet (\Tbl{scannet_scenewise}), semantic segmentation on S3DIS (\Tbl{stanford_scenewise}), and qualitative visualizations of semantic segmentation results on the S3DIS dataset (\Fig{stanford_semseg}), instance segmentation on the ScanNet dataset (\Fig{scannet_insseg}), instance segmentation on the S3DIS dataset (\Fig{stanford_insseg}), and semantic segmentation on the SemanticKITTI dataset (\Fig{kitti_semseg}). We use compressed Res16UNet34C models for all experiments.

\begin{table*}[h]
\centering
\small
\caption{Architectures of Res16UNet variants. We denote a convolution layer with its kernel size, output channel size, and convolution stride size. If a stride side is not specified, it is assumed that there is no stride. All convolution layers except for the last layer have a Batch Normalization and a ReLU layer after them. The layers with the tag "conv\_tr" indicates the transposed convolution layers. We use a square bracket to denote a residual block, with the number of blocks stacked. For instance segmentation, the additional head (offset${}^*$) for offset estimation exists, as depicted in Figure \textcolor{red}{2} (Top) in our main paper.}
  \resizebox{0.85\textwidth}{!}{
  \begin{tabular}{l|c|c|c}
    \toprule
    layer name & Res16UNet14A & Res16UNet18A & Res16UNet34C \\ 
    \midrule
    conv0 & \multicolumn{3}{c}{$3\times3\times3$, 32} \\
    \midrule
    conv1 & \multicolumn{3}{c}{$2\times2\times2$, 32, stride 2} \\
    \midrule
    block1 & $\begin{bmatrix} 3\times3\times3, \text{32} \\ 3\times3\times3, 32 \end{bmatrix} \times 1$ & $\begin{bmatrix} 3\times3\times3, \text{32} \\ 3\times3\times3, 32 \end{bmatrix} \times 2$ & $\begin{bmatrix} 3\times3\times3, \text{32} \\ 3\times3\times3, 32 \end{bmatrix} \times 2$ \\
    \midrule
    conv2 & \multicolumn{3}{c}{$2\times2\times2$, 32, stride 2} \\
    \midrule
    block2 & $\begin{bmatrix} 3\times3\times3, \text{64} \\ 3\times3\times3, \text{64} \end{bmatrix} \times 1$ & $\begin{bmatrix} 3\times3\times3, \text{64} \\ 3\times3\times3, \text{64} \end{bmatrix} \times 2$ & $\begin{bmatrix} 3\times3\times3, \text{64} \\ 3\times3\times3, \text{64} \end{bmatrix} \times 3$ \\
    \midrule
    conv3 & \multicolumn{3}{c}{$2\times2\times2\times$, 64, stride 2} \\
    \midrule
    block3 & $\begin{bmatrix} 3\times3\times3, \text{128} \\ 3\times3\times3, \text{128} \end{bmatrix} \times 1$ & $\begin{bmatrix} 3\times3\times3, \text{128} \\ 3\times3\times3, \text{128} \end{bmatrix} \times 2$ & $\begin{bmatrix} 3\times3\times3, \text{128} \\ 3\times3\times3, \text{128} \end{bmatrix} \times 4$ \\
    \midrule
    conv4 & \multicolumn{3}{c}{$2\times2\times2$, 128, stride 2} \\
    \midrule
    block4 & $\begin{bmatrix} 3\times3\times3, \text{256} \\ 3\times3\times3, \text{256} \end{bmatrix} \times 1$ & $\begin{bmatrix} 3\times3\times3, \text{256} \\ 3\times3\times3, \text{256} \end{bmatrix} \times 2$ & $\begin{bmatrix} 3\times3\times3, \text{256} \\ 3\times3\times3, \text{256} \end{bmatrix} \times 6$ \\
    \midrule
    conv4\_tr & \multicolumn{2}{c|}{$2\times2\times2$, 128, stride 2} & $2\times2\times2$, 256, stride 2 \\
    \midrule
    block5 & $\begin{bmatrix} 3\times3\times3, \text{128} \\ 3\times3\times3, \text{128} \end{bmatrix} \times 1$ & $\begin{bmatrix} 3\times3\times3, \text{128} \\ 3\times3\times3, \text{128} \end{bmatrix} \times 2$ & $\begin{bmatrix} 3\times3\times3, \text{256} \\ 3\times3\times3, \text{256} \end{bmatrix} \times 2$ \\
    \midrule
    conv5\_tr & \multicolumn{3}{c}{$2\times2\times2$, 128, stride 2} \\
    \midrule
    block6 & $\begin{bmatrix} 3\times3\times3, \text{128} \\ 3\times3\times3, \text{128} \end{bmatrix} \times 1$ & $\begin{bmatrix} 3\times3\times3, \text{128} \\ 3\times3\times3, \text{128} \end{bmatrix} \times 2$ & $\begin{bmatrix} 3\times3\times3, \text{128} \\ 3\times3\times3, \text{128} \end{bmatrix} \times 2$ \\
    \midrule
    conv6\_tr & \multicolumn{3}{c}{$2\times2\times2$, 96, stride 2} \\
    \midrule
    block7 & $\begin{bmatrix} 3\times3\times3, \text{96} \\ 3\times3\times3, \text{96} \end{bmatrix} \times 1$ & $\begin{bmatrix} 3\times3\times3, \text{96} \\ 3\times3\times3, \text{96} \end{bmatrix} \times 2$ & $\begin{bmatrix} 3\times3\times3, \text{96} \\ 3\times3\times3, \text{96} \end{bmatrix} \times 2$ \\
    \midrule
    conv7\_tr & \multicolumn{3}{c}{$2\times2\times2$, 96, stride 2} \\
    \midrule
    block8 & $\begin{bmatrix} 3\times3\times3, \text{96} \\ 3\times3\times3, \text{96} \end{bmatrix} \times 1$ & $\begin{bmatrix} 3\times3\times3, \text{96} \\ 3\times3\times3, \text{96} \end{bmatrix} \times 2$ & $\begin{bmatrix} 3\times3\times3, \text{96} \\ 3\times3\times3, \text{96} \end{bmatrix} \times 2$ \\
    \midrule
    final & \multicolumn{3}{c}{$1\times1\times1, N_C$} \\
    \midrule
    offset${}^{*}$ & \multicolumn{3}{c}{$\begin{bmatrix}1\times1\times1,96 \\ 1\times1\times1, 3\end{bmatrix}$} \\
    \midrule
    Number of Params & $8.02\times10^6$ & $15.5\times10^6$ & $37.9\times10^6$ \\ 
    \bottomrule
  \end{tabular}
  }
\label{tbl:architecture}
\end{table*}

\clearpage
\begin{table}[!htb]
\setlength{\tabcolsep}{2pt}
    \caption{Semantic segmentation results on ScanNet validation split}
    \centering
    \small
    \resizebox{.99\textwidth}{!}{
    \begin{tabular}{l|cccccccccccccccccccc|c}
    IoU & bath & bed & bksf & cab & chair & cntr & curt & desk & door & floor & othr & pic & ref & show & sink & sofa & tab & toil & wall & wind & \textbf{mIoU} \\
    \hline
    Res16UNet14A & 85.1 & 78.2 & 79.1 & 61.0 & 89.7 & 57.6 & 66.5 & 60.1 & 58.2 & 95.0 & 57.2 & 31.0 & 44.3 & 55.4 & 62.8 & 82.0 & 71.0 & 92.8 & 83.0 & 60.4 & 68.5 \\
    $\vdash$ 90\% prune & 84.4 & 77.5 & 79.8 & 61.7 & 88.8 & 56.0 & 67.2 & 59.0 & 59.5 & 94.8 & 56.0 & 27.8 & 45.9 & 53.8 & 63.1 & 80.2 & 69.5 & 90.1 & 83.0 & 59.1 & 67.9 \\
    $\vdash$ 95\% prune & 83.0 & 76.0 & 77.9 & 60.4 & 88.3 & 58.0 & 65.5 & 57.6 & 58.1 & 94.8 & 53.1 & 28.6 & 45.9 & 53.7 & 62.9 & 81.1 & 69.2 & 89.5 & 82.0 & 57.8 & 67.2 \\
    $\vdash$ 99\% prune & 69.8 & 70.2 & 73.2 & 50.5 & 83.3 & 44.9 & 60.0 & 52.0 & 49.8 & 94.1 & 43.2 & 19.8 & 36.7 & 47.6 & 46.7 & 75.4 & 62.6 & 77.4 & 78.5 & 50.7 & 59.3 \\
    \hline
    Res16UNet18A & 85.3 & 80.9 & 78.6 & 61.0 & 89.7 & 63.2 & 73.0 & 62.3 & 65.1 & 95.0 & 55.8 & 30.4 & 48.5 & 62.0 & 63.3 & 80.4 & 70.7 & 90.7 & 84.6 & 65.0 & 70.3 \\
    $\vdash$ 90\% prune & 85.2 & 78.9 & 79.7 & 62.1 & 89.2 & 60.5 & 72.5 & 60.6 & 65.2 & 95.0 & 55.5 & 31.6 & 47.0 & 60.3 & 65.2 & 80.1 & 70.9 & 92.3 & 84.5 & 61.9 & 69.9 \\
    $\vdash$ 95\% prune & 82.7 & 78.7 & 77.3 & 60.9 & 89.1 & 59.8 & 71.2 & 62.0 & 63.8 & 94.9 & 55.5 & 29.7 & 49.5 & 58.2 & 64.9 & 79.2 & 71.2 & 91.3 & 84.0 & 61.2 & 69.3 \\
    $\vdash$ 99\% prune & 80.4 & 75.9 & 77.9 & 59.1 & 86.0 & 55.6 & 68.9 & 59.3 & 58.3 & 94.5 & 52.0 & 28.2 & 41.5 & 54.6 & 54.2 & 76.8 & 66.7 & 85.6 & 82.1 & 56.2 & 65.7 \\
    \hline
    Res16UNet34C & 86.1 & 80.0 & 80.1 & 65.0 & 90.4 & 65.4 & 73.4 & 64.3 & 65.6 & 95.0 & 59.3 & 30.1 & 51.4 & 67.1 & 64.5 & 82.2 & 73.0 & 91.2 & 85.0 & 62.5 & 71.6 \\
    $\vdash$ 90\% prune & 86.9 & 79.2 & 78.9 & 64.8 & 90.7 & 63.0 & 72.6 & 61.9 & 63.4 & 95.0 & 57.9 & 29.7 & 50.7 & 68.1 & 66.0 & 81.7 & 72.7 & 91.4 & 84.5 & 61.5 & 71.0 \\
    $\vdash$ 95\% prune & 85.9 & 79.8 & 79.0 & 65.0 & 90.3 & 61.4 & 74.0 & 62.7 & 63.9 & 95.1 & 57.6 & 28.9 & 51.4 & 67.9 & 64.4 & 81.5 & 73.1 & 90.9 & 84.5 & 61.5 & 71.0 \\
    $\vdash$ 99\% prune & 83.8 & 78.2 & 76.9 & 62.8 & 88.9 & 60.2 & 72.1 & 61.6 & 64.7 & 94.9 & 54.8 & 30.1 & 48.9 & 65.6 & 60.8 & 80.2 & 70.1 & 89.5 & 83.5 & 61.7 & 69.5 \\
     
    \multicolumn{21}{c}{\vspace{0.3cm}}\\
    Acc & bath & bed & bksf & cab & chair & cntr & curt & desk & door & floor & othr & pic & ref & show & sink & sofa & tab & toil & wall & wind & \textbf{mAcc} \\
    \hline
    Res16UNet14A & 91.8 & 84.6 & 91.6 & 77.1 & 94.5 & 74.3 & 72.5 & 79.0 & 68.8 & 98.1 & 64.2 & 42.4 & 51.3 & 61.3 & 68.4 & 90.4 & 80.7 & 95.2 & 93.8 & 73.7 & 77.7 \\
    $\vdash$ 90\% prune & 90.1 & 83.7 & 90.1 & 75.7 & 93.8 & 71.1 & 75.2 & 77.6 & 72.0 & 98.0 & 62.3 & 37.8 & 53.3 & 59.0 & 72.5 & 90.3 & 80.9 & 94.0 & 93.7 & 74.7 & 77.3 \\
    $\vdash$ 95\% prune & 87.5 & 83.2 & 89.6 & 75.0 & 93.5 & 73.0 & 74.2 & 78.4 & 70.3 & 98.0 & 59.8 & 37.3 & 54.3 & 60.2 & 72.4 & 90.8 & 79.6 & 93.8 & 93.3 & 73.2 & 76.9 \\
    $\vdash$ 99\% prune & 79.1 & 80.4 & 86.7 & 65.1 & 89.0 & 58.2 & 69.8 & 74.5 & 63.0 & 97.8 & 50.4 & 24.0 & 44.8 & 55.9 & 59.0 & 88.8 & 73.8 & 86.9 & 92.2 & 67.3 & 70.3 \\
    \hline
    Res16UNet18A & 93.1 & 85.5 & 88.9 & 75.4 & 93.9 & 77.2 & 82.7 & 82.5 & 77.0 & 98.1 & 62.0 & 39.4 & 56.4 & 68.4 & 75.7 & 92.0 & 80.3 & 93.8 & 94.7 & 76.4 & 79.7 \\
    $\vdash$ 90\% prune & 92.6 & 85.3 & 88.6 & 76.6 & 93.7 & 74.8 & 80.4 & 81.2 & 77.8 & 98.0 & 61.4 & 38.9 & 53.8 & 65.4 & 75.7 & 91.9 & 80.9 & 94.6 & 94.7 & 74.6 & 79.0 \\
    $\vdash$ 95\% prune & 90.8 & 84.8 & 87.6 & 75.1 & 93.8 & 75.2 & 79.5 & 79.3 & 77.3 & 98.0 & 61.7 & 38.0 & 56.6 & 63.0 & 75.9 & 91.3 & 81.6 & 94.3 & 94.3 & 74.4 & 78.6 \\
    $\vdash$ 99\% prune & 87.9 & 83.0 & 88.8 & 72.6 & 91.2 & 71.3 & 79.7 & 79.9 & 70.7 & 97.8 & 59.7 & 35.1 & 50.2 & 64.1 & 65.7 & 89.9 & 77.4 & 92.9 & 93.6 & 70.8 & 76.1 \\
    \hline
    Res16UNet34C & 92.7 & 85.2 & 91.9 & 78.7 & 94.9 & 77.8 & 78.3 & 87.3 & 79.7 & 98.1 & 64.4 & 36.3 & 58.2 & 73.3 & 75.6 & 91.2 & 79.3 & 95.3 & 94.5 & 76.1 & 80.4 \\
    $\vdash$ 90\% prune & 92.7 & 84.9 & 90.1 & 78.4 & 94.7 & 74.5 & 77.1 & 84.3 & 75.4 & 98.2 & 62.3 & 36.0 & 56.5 & 74.6 & 77.3 & 91.3 & 81.2 & 95.7 & 94.8 & 76.1 & 79.8 \\ 
    $\vdash$ 95\% prune & 92.7 & 85.1 & 90.8 & 77.7 & 94.2 & 74.3 & 79.4 & 83.5 & 75.9 & 98.1 & 63.4 & 35.2 & 56.8 & 73.4 & 75.4 & 91.8 & 82.6 & 94.6 & 94.7 & 75.5 & 79.7 \\
    $\vdash$ 99\% prune & 89.8 & 84.1 & 88.2 & 77.3 & 93.4 & 73.6 & 79.5 & 82.5 & 78.7 & 97.8 & 61.5 & 38.7 & 53.6 & 73.9 & 72.2 & 91.0 & 80.4 & 94.5 & 93.9 & 74.3 & 78.9 \\
    \end{tabular}
    }
    \label{tbl:scannet_scenewise}
\end{table}
\begin{table*}[!htb]
\setlength{\tabcolsep}{2pt}
    \caption{Semantic segmentation results on S3DIS Area 5 Test}
    \centering
    \small
    \resizebox{.99\textwidth}{!}{
    \begin{tabular}{l|ccccccccccccc|c}
    IoU & beam & board & bookcase & ceiling & chair & clutter & column & door & floor & sofa & table & wall & window & \textbf{mIoU} \\
    \hline
    Res16UNet14A & 0.0 & 75.6 & 72.5 & 91.3 & 87.4 & 57.2 & 35.6 & 76.9 & 96.0 & 50.6 & 80.0 & 82.9 & 53.1 & 66.1 \\
    $\vdash$ 90\% prune & 0.0 & 76.0 & 72.2 & 92.3 & 88.7 & 57.6 & 29.2 & 71.0 & 95.8 & 43.9 & 78.9 & 82.8 & 58.2 & 65.1 \\
    $\vdash$ 95\% prune & 0.2 & 74.2 & 71.7 & 92.4 & 87.6 & 57.3 & 30.8 & 69.1 & 96.5 & 43.2 & 79.0 & 83.0 & 57.7 & 64.8 \\
    $\vdash$ 99\% prune & 0.1 & 72.6 & 68.6 & 90.9 & 84.8 & 54.3 & 26.6 & 61.8 & 93.5 & 30.0 & 71.9 & 81.9 & 54.4 & 60.9 \\
    \hline
    Res16UNet18A & 0.0 & 75.8 & 72.9 & 91.5 & 88.1 & 57.4 & 29.6 & 76.2 & 97.1 & 67.0 & 79.7 & 84.5 & 59.1 & 67.6 \\
    $\vdash$ 90\% prune & 0.0 & 77.7 & 73.8 & 90.8 & 89.0 & 57.4 & 26.0 & 76.4 & 97.3 & 54.1 & 79.9 & 84.0 & 57.9 & 66.5 \\
    $\vdash$ 95\% prune & 0.0 & 75.8 & 72.9 & 90.0 & 88.9 & 55.8 & 30.9 & 71.8 & 96.5 & 59.1 & 77.8 & 84.1 & 58.2 & 66.3 \\
    $\vdash$ 99\% prune & 0.0 & 75.2 & 70.3 & 88.4 & 88.4 & 53.5 & 27.9 & 67.8 & 95.3 & 52.6 & 77.5 & 83.5 & 54.5 & 64.2 \\
    \hline
    Res16UNet34C & 0.1 & 79.4 & 71.0 & 92.7 & 88.9 & 57.6 & 40.0 & 73.2 & 95.1 & 72.6 & 78.6 & 84.9 & 60.1 & 68.8 \\
    $\vdash$ 90\% prune & 0.0 & 74.1 & 72.7 & 93.4 & 88.6 & 59.2 & 26.0 & 67.1 & 96.1 & 57.6 & 78.8 & 84.5 & 59.8 & 66.5 \\
    $\vdash$ 95\% prune & 0.0 & 75.5 & 72.4 & 92.6 & 87.6 & 58.1 & 27.4 & 69.7 & 96.0 & 61.1 & 78.6 & 84.5 & 59.5 & 66.4 \\
    $\vdash$ 99\% prune & 0.0 & 73.4 & 72.1 & 91.4 & 88.5 & 56.6 & 31.1 & 66.3 & 95.5 & 60.1 & 76.4 & 84.0 & 58.0 & 65.7 \\
     
    \multicolumn{15}{c}{\vspace{0.3cm}}\\
    Acc & beam & board & bookcase & ceiling & chair & clutter & column & door & floor & sofa & table & wall & window & \textbf{mAcc} \\
    \hline
    Res16UNet14A & 0.0 & 81.8 & 80.7 & 93.7 & 96.1 & 74.2 & 55.7 & 86.6 & 98.3 & 68.7 & 89.6 & 94.0 & 55.0 & 75.0 \\
    $\vdash$ 90\% prune & 0.0 & 82.8 & 80.5 & 95.3 & 96.1 & 76.4 & 34.9 & 79.6 & 97.5 & 48.7 & 89.5 & 94.8 & 60.4 & 72.0 \\
    $\vdash$ 95\% prune & 1.5 & 81.4 & 81.0 & 95.1 & 95.7 & 76.1 & 38.0 & 78.0 & 97.9 & 49.9 & 89.3 & 94.4 & 60.7 & 72.2 \\
    $\vdash$ 99\% prune & 1.1 & 81.2 & 78.3 & 93.3 & 93.5 & 75.2 & 29.5 & 70.6 & 95.6 & 33.8 & 88.7 & 94.9 & 57.3 & 68.7 \\
    \hline
    Res16UNet18A & 0.0 & 84.8 & 82.6 & 93.8 & 96.1 & 73.9 & 33.0 & 87.3 & 98.5 & 72.7 & 89.3 & 95.4 & 62.1 & 74.6 \\
    $\vdash$ 90\% prune & 0.0 & 84.1 & 82.1 & 93.7 & 96.3 & 75.7 & 29.1 & 85.2 & 98.6 & 59.4 & 88.9 & 95.5 & 60.5 & 73.0 \\
    $\vdash$ 95\% prune & 0.0 & 84.4 & 82.0 & 92.7 & 96.3 & 73.8 & 33.5 & 82.4 & 98.1 & 62.9 & 89.1 & 95.7 & 60.7 & 73.2 \\
    $\vdash$ 99\% prune & 0.0 & 83.4 & 80.0 & 90.4 & 95.3 & 76.2 & 31.4 & 78.0 & 97.0 & 56.6 & 88.3 & 95.5 & 56.7 & 71.4 \\
    \hline
    Res16UNet34C & 0.4 & 84.2 & 80.0 & 95.9 & 95.2 & 74.0 & 44.1 & 90.3 & 96.5 & 78.0 & 88.4 & 95.7 & 63.2 & 75.8 \\
    $\vdash$ 90\% prune & 0.0 & 86.0 & 81.8 & 95.8 & 96.2 & 77.9 & 27.7 & 77.8 & 97.6 & 60.7 & 88.5 & 95.8 & 61.3 & 73.1 \\ 
    $\vdash$ 95\% prune & 0.1 & 84.9 & 81.8 & 95.0 & 96.2 & 76.4 & 29.4 & 81.3 & 98.1 & 63.3 & 88.2 & 95.7 & 61.5 & 73.2 \\
    $\vdash$ 99\% prune & 0.0 & 85.3 & 81.0 & 94.3 & 95.6 & 76.7 & 33.6 & 75.6 & 97.2 & 63.3 & 88.0 & 95.3 & 60.0 & 72.8 \\
    \end{tabular}
    }
    \label{tbl:stanford_scenewise}
\end{table*}


\end{document}